%% file: main.tex
\documentclass{article}
\input{math_commands.tex}

\usepackage{hyperref}
\usepackage{url}
\usepackage{graphicx}
\usepackage{subcaption}
\usepackage{multirow}
\usepackage{booktabs}       
\usepackage{wrapfig}
\usepackage{caption}
\usepackage{amsmath}
\usepackage{amssymb}
\usepackage{mathtools}
\usepackage{amsthm}
\usepackage{hyperref}

\usepackage[accepted]{icml2023}

\usepackage{amsmath}
\usepackage{amssymb}
\usepackage{mathtools}
\usepackage{amsthm}

\usepackage[capitalize,noabbrev]{cleveref}

\theoremstyle{plain}
\newtheorem{theorem}{Theorem}[section]

\newtheorem{lemma}[theorem]{Lemma}

\theoremstyle{definition}

\theoremstyle{remark}

\usepackage[textsize=tiny]{todonotes}

\icmltitlerunning{Why Random Pruning Is All We Need to Start Sparse}

\begin{document}

\twocolumn[
\icmltitle{Why Random Pruning Is All We Need to Start Sparse}

\icmlsetsymbol{equal}{*}

\begin{icmlauthorlist}
\icmlauthor{Advait Gadhikar}{yyy}
\icmlauthor{Sohom Mukherjee}{yyy}
\icmlauthor{Rebekka Burkholz}{yyy}
\end{icmlauthorlist}

\icmlaffiliation{yyy}{CISPA Helmholtz Center for Information Security, Saarbrücken, Germany}

\icmlcorrespondingauthor{Advait Gadhikar}{advait.gadhikar@cispa.de}

\icmlkeywords{Lottery Ticket Hypothesis, Random Pruning}

\vskip 0.3in
]



\printAffiliationsAndNotice{\icmlEqualContribution} 

\begin{abstract}
Random masks define surprisingly effective sparse neural network models, as has been shown empirically.
The resulting sparse networks can often compete with dense architectures and state-of-the-art lottery ticket pruning algorithms, even though they do not rely on computationally expensive prune-train iterations and can be drawn initially without significant computational overhead. 
We offer a theoretical explanation of how random masks can approximate arbitrary target networks if they are wider by a logarithmic factor in the inverse sparsity $1 / \log(1/\text{sparsity})$. 
This overparameterization factor is necessary at least for 3-layer random networks, which elucidates the observed degrading performance of random networks at higher sparsity.
At moderate to high sparsity levels, however, our results imply that sparser networks are contained within random source networks so that any dense-to-sparse training scheme can be turned into a computationally more efficient sparse to sparse one by constraining the search to a fixed random mask.
We demonstrate the feasibility of this approach in experiments for different pruning methods and propose particularly effective choices of initial layer-wise sparsity ratios of the random source network.
As a special case, we show theoretically and experimentally that random source networks also contain strong lottery tickets. 
Our code is available at \url{https://github.com/RelationalML/sparse_to_sparse}.
\end{abstract}

\input{intro.tex}

\input{theory.tex}

\input{expt.tex}


\bibliography{example_paper}
\bibliographystyle{icml2023}

\newpage
\appendix
\onecolumn
\input{app.tex}


\end{document}

%% file: math_commands.tex
\usepackage{amsmath,amsfonts,bm}

\def\eqref#1{equation~\ref{#1}}

\def\1{\bm{1}}

\DeclareMathAlphabet{\mathsfit}{\encodingdefault}{\sfdefault}{m}{sl}
\SetMathAlphabet{\mathsfit}{bold}{\encodingdefault}{\sfdefault}{bx}{n}



%% file: intro.tex
\section{Introduction}


The impressive breakthroughs achieved by deep learning have largely been attributed to the extensive overparametrization of deep neural networks, as it seems to have multiple benefits for their representational power and optimization \citep{belkin-overparam}.
The resulting trend towards ever larger models and datasets, however, imposes increasing computational and energy costs that are difficult to meet. 
This raises the question: Is this high degree of overparameterization truly necessary?

Training general small-scale or sparse deep neural network architectures from scratch remains a challenge for standard initialization schemes \cite{pruning-filters-for-efficient-cnns,han-efficient-nns}. 
However, \cite{weak-lth} have recently demonstrated that there exist sparse architectures that can be trained to solve standard benchmark problems competitively. 
According to their Lottery Ticket Hypothesis (LTH), dense randomly initialized networks contain subnetworks that can be trained in isolation to a test accuracy that is comparable with the one of the original dense network. 
Such subnetworks, the lottery tickets (LTs), have since been obtained by pruning algorithms that require computationally expensive pruning-retraining iterations \citep{weak-lth,synflow} or mask learning procedures \cite{continuous-sparsification,rare-gems}.
While these can lead to computational gains at training and inference time and reduce memory requirements \citep{optimal-brain-surgeon,han-efficient-nns}, the real goal remains to identify sparse trainable architectures before training, as this could lead to significant computational savings. 
Yet, contemporary pruning at initialization approaches \citep{snip,grasp,synflow,planting-jonas,missing-mark} achieve less competitive performance.
For that reason it is so remarkable that even iterative state-of-the-art approaches struggle to outperform a simple, computationally cheap, and data independent alternative: random pruning at initialization \citep{sanity-check-1}.
\citet{random-pruning-vita} have provided systematic experimental evidence for its 'unreasonable` effectiveness in multiple settings, including complex, large scale architectures and data.

We explain theoretically why they can be effective by proving that a randomly masked network can approximate an arbitrary target network if it is wider by a logarithmic factor in its sparsity $1 / \log(1/\text{sparsity})$.
By deriving a lower bound on the required width of a random 1-hidden layer network, we further show that this degree of overparameterization is necessary in general. 
This implies that sparse random networks have the universal function approximation property like dense networks and are at least as expressive as potential target networks. 
However, it also highlights the limitations of random pruning in case of extremely high sparsities, as the width requirement scales then approximately as $1 / \log(1/\text{sparsity}) \approx 1/(1-\text{sparsity})$ (see also Fig.~\ref{fig:er-width} for an example).
In practice, we observe a similar degradation in performance for high sparsity levels.

Even for moderate to high sparsities, the randomness of the connections result in a considerable number of excess weights that are not needed for the representation of a target network. 
This insight suggests that, on the one hand, additional pruning could further enhance the sparsity of the resulting neural network structure, as random masks are likely not optimally sparse. 
On the other hand, any dense-to-sparse training approach would not need to start from a dense network but could also start training from a sparser random network and thus be turned into a sparse-to-sparse learning method. 
\begin{figure}[h!]
\centering
\includegraphics[width=0.5\textwidth]{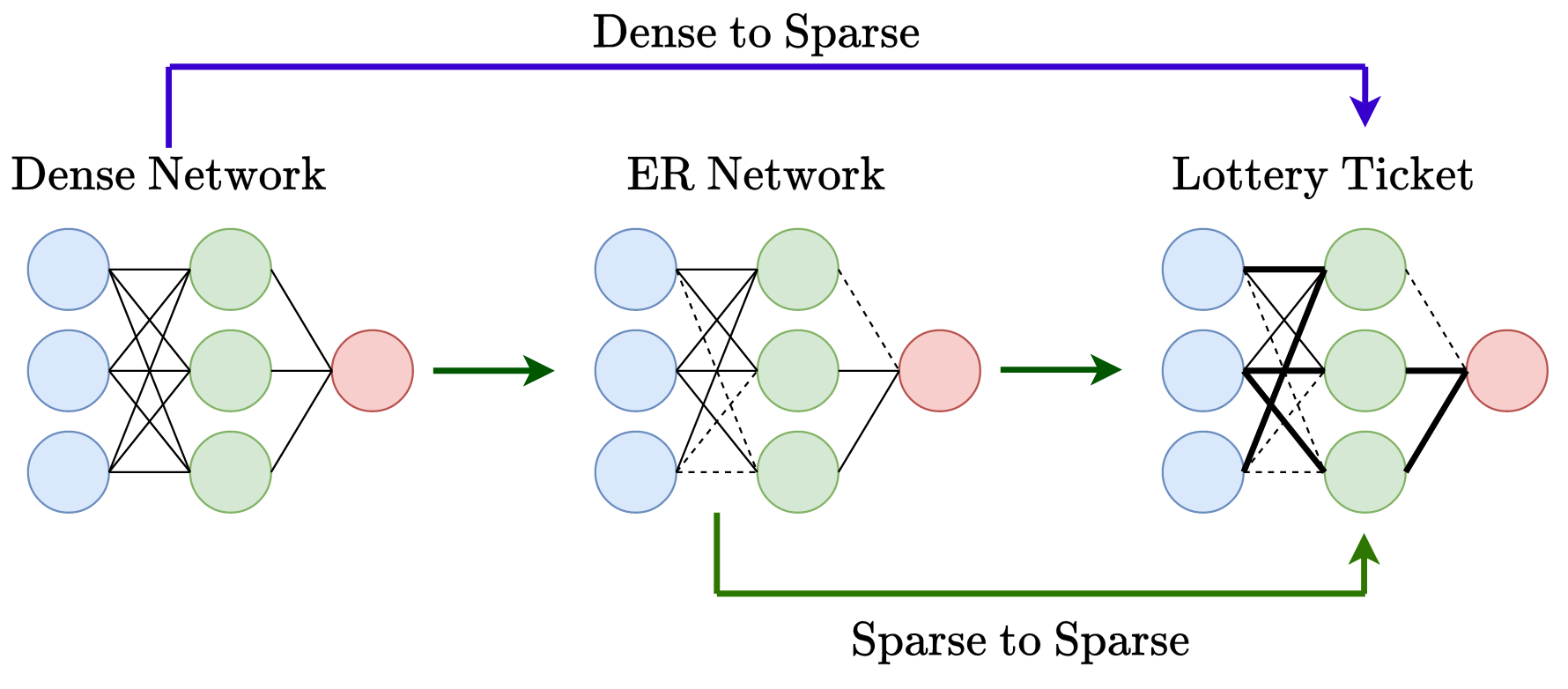}
\caption{\textit{Sparse training with randomly masked (ER) networks:}
A visual representation of the main implication of our theory - sparse to sparse training can be effective by starting from a randomly masked (ER) network.
}\label{fig:er-explained}
\end{figure}
The main idea is visualized in Fig.~\ref{fig:er-explained} and verified in extensive experiments with different lottery ticket pruning and continuous sparsification approaches. 
Our main results could also be interpreted as theoretical justification for Dynamic Sparse Training (DST) \citep{evci-rigl, granet, deep-rewiring}, which prunes random networks of moderate sparsity.
However, it further relies on edge rewiring steps that sometimes require the computation of gradients of the corresponding dense network \citep{evci-rigl}.
Our derived limitations of random pruning indicate that this rewiring might be necessary at extreme sparsities but likely not for moderately sparse random starting points, as we also highlight in additional experiments.

As a special case of the main idea to prune random networks, we also consider strong lottery tickets (SLTs) \cite{deconstructing-lt,ramanujan-slth}.
These are subnetworks of large, randomly initialized source networks, which do not require any further training after pruning.
Theoretical \citep{malach-proof-lth,pensia-subset-sum,nonzerobiases,cnnexist,convexist,depthexist,uniExist} as well as empirical \citet{ramanujan-slth,deconstructing-lt,diffenderfer2021multiprize,sreenivasan22a} existence proofs so far have solely focused on pruning dense source networks.
We highlight the potential for computational resource savings in the search for SLTs by proving their existence within sparse random networks instead. 
The main component of our results is Lemma~\ref{lemma:subset}, which extends subset sum approximations to the sparse random graph setting. 
This enables the direct transfer of most SLT existence results for different architectures and activation functions to sparse source networks.
Furthermore, we modify the algorithm edge-popup (EP) \cite{ramanujan-slth} to find SLTs accordingly which leads to the first sparse-to-sparse pruning approach for SLTs, up to our knowledge.
We demonstrate in experiments that starting even at sparsities as high as 0.8 does not hamper the overall performance of EP.

Note that our general theory applies to any layerwise sparsity ratios of the random source network and we validate this fact in various experiments on standard benchmark image data and commonly used neural network architectures, complementing results by \citet{random-pruning-vita} for additional choices of sparsity ratios.
Our two proposals, balanced and pyramidal sparsity ratios, seem to perform competitively across multiple settings, especially, at higher sparsity regimes.

\paragraph{Contributions}
\begin{enumerate}

\item We prove that randomly pruned random networks are sufficiently expressive and can approximate an arbitrary target network if they are wider by a factor of $1/\log(1/\text{sparsity})$. This overparametrization factor is necessary in general, as our lower bound for univariate target networks indicates.

\item Inspired by our proofs, we empirically demonstrate that, without significant loss in performance, starting any dense-to-sparse training scheme can be translated into a sparse-to-sparse one by starting from a random source network instead of a dense one.

\item As a special case, we also prove the existence of Strong Lottery Tickets (SLTs) within sparse random source networks, if the source network is wider than a target by a factor  $1/\log(1/\text{sparsity})$. 
Our modification of the edge-popup (EP) algorithm \citep{ramanujan-slth} leads to the first sparse-to-spare SLT pruning method, which validates our theory and highlights potential for computational savings.

\item To demonstrate that our theory applies to various choices of sparsity ratios, we introduce two additional proposals that outperform state-of-the-art ones on multiple benchmarks and are thus promising candidates for starting points of sparse-to-sparse learning schemes.

\end{enumerate}

\subsection{Related Work}
Algorithms to prune neural networks for unstructured sparsity can be broadly categorized into two groups, pruning after training and pruning before (or during) training. 
The first group of algorithms that prune after training are effective in speeding up inference, but they still rely on a computationally expensive training procedure \citep{optimal-brain-surgeon,optimal-brain-damage,molchanov-cnn-inference,layerwise-brain-surgeon,combinatorial-brain-surgeon}. 
The second group of algorithms prune at initialization \citep{snip,grasp,synflow,rare-gems,de2020progressive} or follow a computationally expensive cycle of pruning and retraining for multiple iterations \citep{gale-sparsity,continuous-sparsification,early-bird-tickets,weak-lth,weight-rewinding-frankle}. 
These methods find trainable subnetworks also known as Lottery Tickets \citep{weak-lth}.
Single shot pruning approaches are computationally cheaper but are susceptible to problems like layer collapse which render the pruned network untrainable \citep{snip,grasp}. 
\citet{synflow} address this issue by preserving flow in the network through their scoring mechanism. 
The best performing sparse networks are still obtained by expensive iterative pruning methods like Iterative Magnitude Pruning (IMP), Iterative Synflow \citep{weak-lth,planting-jonas} or continuous sparsification methods \citep{rare-gems,continuous-sparsification, kusupati2020soft, louizos2018learning}.

However, \citet{sanity-check-1} found that randomly pruned masks can outperform expensive iterative pruning strategies in different situations.
Inspired by this finding, \citet{are-wider-nets-better,provable-benefits-overparam} have hypothesized that sparse overparameterized networks are more effective than smaller networks with the same number of parameters.
\citet{random-pruning-vita} have further demonstrated the competitiveness of random masks for different data independent choices of layerwise sparsity ratios across a wide range of neural network architectures and datasets, including complex ones. 
Our analysis identifies the conditions under which the effectiveness of random masks is reasonable.
We show that a sparse random source network can approximate a target network if it is wider by a factor proportional to the inverse log sparsity.
Complementing experiments by \citet{random-pruning-vita}, we highlight that random masks are competitive for various choices of layerwise sparsity ratios.
However, we also show that their randomness also likely induces potential for further pruning. 

We build on the lottery ticket existence theory \citep{malach-proof-lth,pensia-subset-sum,orseau-log-slth,nonzerobiases,uniExist,depthexist,equivariantLT}  to prove that sparse random source networks actually contain strong lottery tickets (SLTs) if their width exceeds a value that is proportional to the width of a target network.  
This theory has been inspired by experimental evidence for SLTs \citep{ramanujan-slth,deconstructing-lt,diffenderfer2021multiprize,sreenivasan22a}. 
The underlying algorithm edge-popup \citep{ramanujan-slth} finds SLTs by training scores for each parameter of the dense source network and is thus computationally as expensive as dense training. 
We show that training smaller random sparse source networks is sufficient, thus, reducing effectively the computational requirements for finding SLTs.

However, our theory suggests that random ER networks face a fundamental limitation at extreme sparsities, as the overparameterization factor scales in this regime as $1/\log(1 /(\text{sparsity})) \approx 1/(1-\text{sparsity})$. 
This shortcoming could be potentially addressed by targeted rewiring of random edges with Dynamical Sparse Training (DST) that starts pruning from an ER network \citep{granet,set-mocanu, yuan2021mest}. 
So far, sparse-to-sparse training methods like \citet{evci-rigl, sparse-momentum-dettmers} still require dense gradients for there edge rewiring operation. 
\citet{zhou2021efficient} obtain sparse training by estimating a sparse gradient using two forward passes.
We empirically show that in light of the expressive power of random networks, we can also achieve sparse-to-sparse training by simply constraining any pruning method or gradient to a fixed initial sparse random mask.

%% file: theory.tex
\section{Expressiveness of Random Networks}
\label{theory}
Our theoretical investigations of the next section have the purpose to explain why the effectiveness of random networks is reasonable given their high expressive power.
We show that we can approximate any target network with the help of a random network, provided that it is wider by a logarithmic factor in the inverse sparsity.
First, the only constraint that we face in our explicit construction of a representative subnetwork is that edges are randomly available or unavailable. But we can choose the remaining network parameters, i.e., the weights and biases, in such a way that we can optimally represent a target network.
As common in results on expressiveness and representational power, we make statements about the existence of such parameters, not necessarily, if they can be found algorithmically.
In practice, the parameters would usually be identified by standard neural network training or prune-train iterations.
Our experiments validate that this is actually feasible in addition to plenty of other experimental evidence \citep{sanity-check-1,sanity-check-jackpot,random-pruning-vita}.
Second, we prove the existence of strong lottery tickets (SLTs), which assumes that we have to approximate the target parameters by pruning the sparse random source network.
Up to our knowledge, we are the first to provide experimental and theoretical evidence for the feasibility of this case. 

\textbf{Background, Notation, and Proof-Setup}
Let $\bm{x} = (x_1, x_2, .., x_d) \in {[a_1,b_1]}^d$ be a bounded $d$-dimensional input vector, where $a_1, b_1 \in \mathbb{R}$ with $a_1 < b_1$. 
$f$: ${[a_1,b_1]}^d \rightarrow \mathbb{R}^{n_L}$ is a fully-connected feed forward neural network with architecture $\left(n_0, n_1, .., n_L\right)$, i.e., depth $L$ and $n_l$ neurons in Layer $l$. 
Every layer $l \in \{1, 2, .., L\}$ computes neuron states $\bm{x}^{(l)} = \phi\left(\bm{h^{(l)}}\right), \bm{h^{(l)}} = \bm{W}^{(l-1)}\bm{x}^{(l-1)} + \bm{b}^{(l-1)}$. 
$\bm{h^{(l)}}$ is called the pre-activation, $\bm{W}^{(l)} \in \mathbb{R}^{n_l \times n_{l-1}}$ is the weight matrix and $\bm{b}^{(l)}$ is the bias vector. 
We also write $f(\bm{x}; \theta)$ to emphasize the dependence of the neural network on its parameters $\theta = (\bm{W}^{(l)}, \bm{b}^{(l)})^L_{l=1}$. 
For simplicity, we restrict ourselves to the common ReLU $\phi(x) = \max\{x, 0\}$ activation function, but most of our results can be easily extended to more general activation functions as in \citep{depthexist,convexist}.
In addition to fully-connected layers, we also consider convolutional layers.
For a convenient notation, without loss of generality, we flatten the weight tensors so that $\bm{W}_T^{(l)} \in \mathbb{R}^{c_l \times c_{l-1} \times k_l}$ where $c_l, c_{l-1}, k_l$ are the output channels, input channels and filter dimension respectively.
For instance, a $2$-dimensional convolution on image data would result in $k_l = k'_{1,l} k'_{2,l}$, where $k'_{1,l}, k'_{2,l}$ define the filter size.

We distinguish three kinds of neural networks, a target network $f_T$, a source network $f_S$, and a subnetwork $f_P$ of $f_S$.
$f_T$ is approximated or exactly represented by $f_P$, which is obtained by masking the parameters of the source $f_S$.
$f_S$ is said to contain a SLT if this subnetwork does not require further training after obtaining the mask (by pruning).
We assume that $f_T$ has depth $L$ and parameters $\left(\bm{W}_T^{(l)}, \bm{b}_T^{(l)}, n_{T,l}, m_{T,l}\right)$ are the weight, bias, number of neurons and number of nonzero parameters of the weight matrix in Layer $l \in \{1, 2, .., L\}$. 
Note that this implies $m_{l} \leq n_{l}n_{l-1}$. 
Similarly, $f_S$ has depth $L+1$ with parameters $\left(\bm{W}_S^{(l)}, \bm{b}_S^{(l)}, n_{S,l}, m_{S,l}\right)^L_{l=0}$. 
Note that $l$ ranges from $0$ to $L$ for the source network, while it only ranges from $1$ to $L$ for the target network.
The extra source network layer $l=0$ accounts for an extra layer that we need in our construction to prove existence.

\textbf{ER Networks}
Even though common, the terminology 'random network` is imprecise with respect to the random distribution from which a graph is drawn.
In line with general graph theory, we therefore use the term Erd\"os-R\'enyi (ER) \cite{erdos1960} network in the following.
An ER neural network $f_{\text{ER}} \in \text{ER}(\mathbf{p})$ is characterized by layerwise sparsity ratios $p_l$. 
An ER source $f_{\text{ER}}$ is defined as a subnetwork of a complete source network using a binary mask $\bm{S}^{(l)}_{\text{ER}} \in \{0,1\}^{n_l \times n_{l-1}}$ or $\bm{S}^{(l)}_{\text{ER}} \in \{0,1\}^{n_l \times n_{l-1} \times k_l}$ for every layer. 
The mask entries are drawn from independent Bernoulli distributions with layerwise success probability $p_l > 0$, i.e., $s^{(l)}_{ij, \text{ER}} \sim \text{Ber}(p_l)$.
The random pruning is performed initially with negligible computational overhead and the mask stays fixed during training. 
Note that $p_l$ is also the expected density of that layer. 
The overall expected density of the network is given as $p = \frac{\sum_l m_l p_l}{\sum_k m_k} = 1 - \text{sparsity}$. 
In case of uniform sparsity, $p_l = p$, we also write $\text{ER}(p)$ instead of $\text{ER}(\mathbf{p})$.
An ER network is defined as $f_{\text{ER}} = f_S(\bm{x}; \bm{W}\cdot\bm{S}_{\text{ER}})$. 
Different to conventional SLT existence proofs \citep{ramanujan-slth}, we refer to $f_{\text{ER}} \in \text{ER}(\mathbf{p})$ as the source network, and show that the SLT is contained within this ER network. 
The SLT is then defined by the mask $S_{\text{P}}$, which is a subnetwork of $S_{\text{ER}}$, i.e., a zero entry $s_{ij, \text{ER}} = 0$ implies also a zero in $s_{ij, \text{P}} = 0$, but the converse is not true. 
We skip the subscripts if the nature of the mask is clear from the context. 
In the following analysis of expressiveness in ER networks, we continue to use of $S_{ER}$ and $S_{P}$ to denote a random ER source network and a sparse subnetwork within the ER network respectively.

\textbf{Sparsity Ratios}
There are plenty of reasonable choices for the layerwise sparsity ratios and thus ER probabilities $p_l$.
Our theory applies to all of them.
The optimal choice for a given source network architecture depends on the target network and thus the solution to a learning problem, which is usually unknown a-priori in practice.
To demonstrate that our theory holds for different approaches, we investigate the following layerwise sparsity ratios in experiments. 
The simplest baseline is a globally \textit{uniform} choice $p_l=p$.
\citet{random-pruning-vita} have compared this choice in extensive experiments with their main proposal, ERK, which assigns $p_l \propto \frac{n_{in} + n_{out}}{n_{in}n_{out}}$ to a linear and $p_l \propto \frac{c_{l} + c_{l-1} + k_l}{c_{l}c_{l-1}k_l}$ \citep{mocanu-evo-train} to a convolutional layer.
In addition, we propose a \textit{pyramidal} and \textit{balanced} approach, which are visualized in Appendix~\ref{app:sparratio}.

\textit{Pyramidal}: This method emulates a property of pruned networks that are obtained by IMP \citep{weak-lth} i.e. the layer densities decay with increasing depth of the network. 
For a network of depth $L$, we use $p_l = \left(p_1\right)^l, p_l \in \left(0,1\right)$ so that $\frac{\sum_{l=1}^{l=L} p_l m_l}{\sum_{l=1}^{l=L}m_l} = p$. 
Given the architecture, we use a polynomial equation solver \cite{numpy} to obtain $p_1$ for the first layer such that $p_1 \in \left(0,1\right)$.

\textit{Balanced}: The second layerwise sparsity method aims to maintain the same number of parameters in every layer for a given network sparsity $p$ and source network architecture. 
Each neuron has a similar in- and out-degree on average.
Every layer has $x = \frac{p}{L}\sum_{l=1}^{l=L}m_l$ nonzero parameters. 
Such an ER network can be realized with $p_l = x / m_l$. 
In case that $x \geq m_l$, we set $p_l=1$.

\subsection{General Expressiveness of ER Networks}

Our main goal in this section is to derive probabilistic statements about the existence of edges in an ER source network that enable us to approximate a given target network.
As every connection in the source network only exists with a probability $p_l$, for each target weight, we need to create multiple candidate edges, of which at least one is nonzero with high enough probability.
This can be achieved by ensuring that each target edge has multiple potential starting points in the ER source network.
Our construction realizes this idea with multiple copies of each neuron in a layer.
The required number of neuron copies depends on the sparsity of the ER source network and introduces an overparametrization factor pertaining to the width of the network.
To create multiple copies of input neurons as well, our construction relies on one additional layer in the source network in comparison with a target network, as visualized in Fig.~ \ref{fig:lt-schematic} in the Appendix.
We first explain the construction for a single target layer and extend it afterwards to deeper architectures.

\paragraph{Single Hidden Layer Targets}
We start with constructing a single hidden layer fully-connected target network with a subnetwork of a random ER source network that consists of one more layer. 
Our proof strategy is visually explained by Fig.~\ref{fig:lt-schematic} in the Appendix.  
The following theorem states the precise width requirement that our construction requires.
\begin{theorem}[Single Hidden Layer Target Construction]
\label{thm:existence-single}
Assume that a single hidden-layer fully-connected target network $f_T(\bm{x}) = \bm{W}_T^{(2)}\phi(\bm{W}_T^{(1)}\bm{x} + \bm{b}_T^{(1)}) + \bm{b}_T^{(2)}$, an allowed failure probability $\delta \in (0,1)$, source densities $\mathbf{p}$ and a $2$-layer ER source network $f_{S} \in \text{ER}(\mathbf{p})$ with widths $n_{S,0} = q_0d, n_{S,1} = q_1n_{T,1}, n_{S,2} = q_2n_{T,2}$ are given. 
If
\begin{align*}
   q_{0} \geq \frac{1}{\log(1 / (1-p_1))}\log\left(\frac{2 m_{T,1}q_1}{\delta}\right),  \\
   q_{1} \geq \frac{1}{\log(1 / (1-p_2))}\log\left(\frac{2 m_{T,2}}{\delta}\right) \text{ and } q_{2} = 1
\end{align*}
then with probability $1-\delta$, the random source network $f_S$ contains a subnetwork $\bm{S}_{P}$ such that $f_S(\bm{x},\bm{W}\cdot\bm{S}_{P}) = f_T$.
\end{theorem}

\textit{Proof Outline}: 

The key idea is to create multiple copies (blocks in Fig.~\ref{fig:lt-schematic}~(b) in the Appendix) in the source network for each target neuron such that every target link is realized by pointing to at least one of these copies in the ER source.
To create multiple candidates of input neurons, we create an univariate first layer in the source network as explained in Fig.~\ref{fig:lt-schematic}. 
In the appendix, we derive the corresponding weight and bias parameters of the source network so that it can represent the target network exactly.
Naturally, many of the available links will receive zero weights if they are not needed in the specific construction but are required for a high enough probability that at least one weight can be set to nonzero.
Our main task in the proof is to estimate the probability that we can find representatives of all target links in the ER source network, i.e., every neuron in Layer $l=1$ has at least one edge to every block in $l=0$ of size $q_0$, as shown in Fig.~\ref{fig:lt-schematic}~(b). 
This probability is given by $\left(1- (1-p_1)^{q_0} \right)^{m_{T,1}  q_1}$.
For the second layer, we repeat a similar argument to bound the probability $\left(1- (1-p_2)^{q_1} \right)^{m_{T,2}}$ with $q_2 = 1$, since we do not require multiple copies of the output neurons.
Bounding this probability by $1-\delta$ completes the proof, as detailed in Appendix~\ref{proof-single}.

\textbf{Deep Target Networks}
Theorem \ref{thm:existence-single} shows that $q_0$ and $q_1$ depend on $1/\log(1 / \text{sparsity})$. 
We now generalize the idea to create multiple copies of target neurons in every layer to a fully connected network of depth $L$ (proofs are in Appendix \ref{proof-general}) and convolutional networks of depth $L$ as stated in Appendix \ref{app:proof-conv}, which yields a similar result as above. 
The additional challenge of the extension is to handle the dependencies of layers, as the construction of every layer needs to be feasible.
\begin{theorem}[ER networks can represent $L$-layer target networks.]
\label{thm:existence}
Given a fully-connected target network $f_T$ of depth $L$, $\delta \in (0,1)$, source densities $\mathbf{p}$ and a $L+1$-layer ER source network $f_{S} \in \text{ER}(\mathbf{p})$ with widths $n_{S,0} = q_0d$ and $n_{S,l} = q_l n_{T,l}, l \in \{1, 2, .., L\}$, where
\begin{align*}
       q_{l} \geq \frac{1}{\log(1 / (1-p_{l+1}))}\log\left(\frac{L m_{T,l+1} q_{l+1} }{\delta}\right) \\
       \text{for } l \in \{0, 1, .., L-1\} \text{ and } q_L = 1,
    \end{align*}
then with probability $1-\delta$ the random source network $f_S$ contains a subnetwork $\bm{S}_{P}$ such that $f_S(\bm{x},\bm{W}\cdot\bm{S}_{P}) = f_T$.
\end{theorem}

\begin{figure}[h!]
    \centering
    
    \includegraphics[height=3.6cm, width=0.3\textwidth]{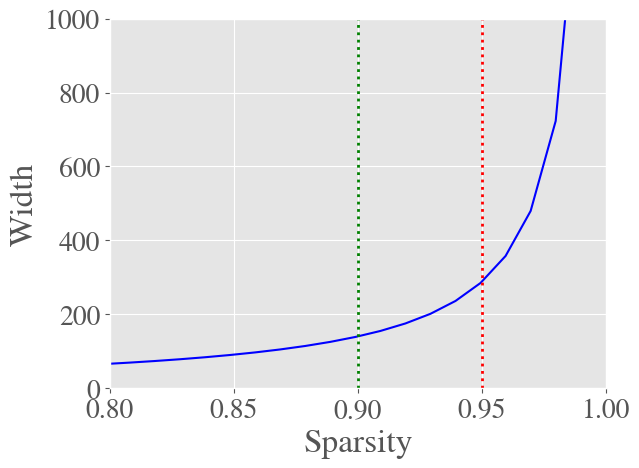}
    
    \caption{\textit{Overparametrization in ER Networks} For a single hidden-layer target network with width $128$ in the hidden layer and $10$ in the output layer, the figure shows the required width of the first layer ($l=1$) of the source ER network as per Theorem \ref{thm:existence-single} with a confidence of $1-\delta = 0.999$.
    The required width increases moderately upto sparsity $0.9$ and drastically after $0.95$.}
    \label{fig:er-width}
\end{figure}

\textbf{Lower Bound on Overparameterization}
While our existence results prove that ER networks have the universal function approximation property like dense neural networks, in order to achieve that, our construction requests a considerable amount of overparametrization in comparison with a dense target network. 
In particularly extremely sparse ER networks seem to face a natural limitation, since for sparsities $1-p \geq 0.9$, the overparameterization factor scales approximately as $1/\log(1 / (1-p)) \approx 1/p$.
Fig.~\ref{fig:er-width} visualizes how this scaling becomes problematic for increasing sparsity. 
The next theorem establishes that, unfortunately, we cannot expect to get around this $1/\log(1 / (1-p_l))$ limitation.
\begin{theorem}[Lower bound on Overparametrization in ER Networks]
\label{thm:lower-bound}
There exist univariate target networks $f_T(\bm{x}) = \phi(\bm{w}^T_T \bm{x} + b_T)$ that cannot be represented by a random $1$-hidden-layer ER source network $f_{S} \in \text{ER}(p)$ with probability at least $1-\delta$, if its width is $n_{S,1} < \frac{1}{\log(1 / (1-p))} \log\left( \frac{1}{1-(1-\delta)^{1/d}}\right)$.
\end{theorem}
 See Fig.~\ref{fig:lower_bound} and App.~\ref{app:lower-bound} for the complete proof.

\textbf{Theoretical Insights}
We have shown that ER networks provably contain subnetworks that can represent general target networks if they are wider by a factor $1 / \log(1 / (1-p_{l}))$. 
This overparameterization factor is necessary and limits the utility of random masks alone to obtain extremely sparse neural network architectures.
However, their high expressiveness make them promising and computationally cheap starting points for further pruning and more general sparsification approaches. 

Inspired by this insight, in the next section, we explore the idea to start pruning from ER source networks in the context of SLTs. 
The first question that we ask is: How much wider do random source networks need to be in order to contain SLTs?

\subsection{Existence of Strong Lottery Tickets}
\label{sec:slts}

Most SLT existence proofs that derive a logarithmic lower bound on the overparametrization factor of the source network \citep{pensia-subset-sum,uniExist,convexist,cnnexist,depthexist,equivariantLT} solve multiple subset sum approximation problems \citep{lueker-subsetsum}. 
For every target parameter $z$, they identify some random parameters of the source network $X_1, ..., X_n$, a subset of which can approximate $z$.
In case of an ER source network, $1-p$ random connections are missing in comparison with a dense source network.
These missing connections also reduce the amount of available source parameters $X_1, ..., X_n$.
To take this into account, we modify the corresponding subset sum approximations according to the following lemma.
\begin{lemma}[Subset sum approximation in ER Networks]
Let $X_1, ..., X_{n}$ be independent, uniformly distributed random variables so that $X_i \sim U([-1,1])$ and $M_1, ..., M_n$ be independent, Bernoulli distributed random variables so that $M_i \sim \text{Ber}(p)$ for a $p>0$.
Let $\epsilon, \delta \in (0,1)$ be given.
Then for any $z \in [-1,1]$ there exists a subset $I \subset [n]$ so that with probability at least $1-\delta$ we have $|z - \sum_{i \in I} M_i X_i| \leq \epsilon$ if
\begin{align}
    n \geq C \frac{1}{\log(1 / (1 - p))} \log\left(\frac{1}{ \min\left(\delta, \epsilon\right)} \right).
\end{align}
\end{lemma}\label{lemma:subset}
The proof is given in App.~\ref{proof-slts} and utilizes the original subset sum approximation result for random subsets of the base set $X_1, ..., X_{n}$. 
In addition, it solves the challenge to combine the involved constants respecting the probability distribution of the random subsets.
For simplicity, we have formulated it for uniform random variables and target parameters $z \in [-1,1]$ but it could be easily extended to random variables that contain a uniform distribution (like normal distributions) and generally bounded targets as in Corollary 7 in \citep{uniExist}.

In comparison with the original subset sum approximation result, we need a base set that is larger by a factor $1/\log(1 / (1 - p))$. 
This is exactly the factor by which we can modify contemporary SLT existence results to transfer to ER source networks and it is also the same factor that we derived in the previous section on expressiveness results.
However, we require in general a higher overparameterization to accommodate subset sum approximations.

The advantage of the formulation of the above lemma is that it allows to transfer general SLT existence results to the ER source setting in a straight forward way.
By replacing the subset sum approximation construction with Lemma~\ref{lemma:subset}, we can thus show SLT existence for fully-connected \citep{pensia-subset-sum,depthexist}, convolutional \citep{uniExist,convexist,cnnexist}, and residual ER networks \citep{convexist}, or random GNNs \citep{equivariantLT}. 
 To give an example for the effective use of this lemma and discuss the general transfer strategy, we explicitly extend the SLT existence results by \citet{depthexist} for fully-connected networks to ER source networks. 
%
We thus show that pruning a random source network of depth $L+1$ with widths larger than a logarithmic factor can approximate any target network of depth $L$ with a given probability $1-\delta$. 

\begin{theorem}[Existence of SLTs in ER Networks]\label{thm:slt} 
Let $\epsilon, \delta \in (0,1)$, a target network $f_T$ of depth $L$, an $\text{ER}(\mathbf{p})$ source network $f_S$ of depth $L+1$ with edge probabilities $p_l$ in each layer $l$ and iid initial parameters $\bm{\theta}$ with $w_{ij}^{(l)} \sim U([-1,1]), b_i^{(l)} \sim U([-1,1])$ be given. 
Then with probability at least $1- \delta$, there exists a mask $\bm{S}_{P}$ so that each target output component $i$ is approximated as $\max_{\bm{x}\in \mathcal{D}}\|f_{T,i}(\bm{x}) - f_{S,i}(\bm{x}; \bm{W}_{S}\cdot\bm{S}_{P} )\| \leq \epsilon $ if 
\begin{align*}
    n_{S,l} \geq C\frac{n_{T,l}}{\log\left(1/(1-p_{l+1})\right)}\log \left(\frac{1}{\min\{\epsilon_{l}, \delta/\rho\}}\right)
\end{align*}
for $\rho = \frac{CN_{T}^{1+\gamma}}{\log(1 / (1-\min_l p_l))^{1+\gamma}} \log(1 / \min\{\min_l\epsilon_l, \delta\})$, $l \geq 1$ for any $\gamma \geq 0$, and where $\epsilon_{l} = g(\epsilon, f_T)$ is defined in App.~\ref{proof-slts}. We also require $n_{S,0} \geq C d 1/\log(1/(1-p_1))\log\left(\frac{1}{\min\{\epsilon_1, \delta / \rho\}}\right)$, where $C>0$ denotes a generic constant that is independent of $n_{T,l}$, $L$, $p_l$, $\delta$, and $\epsilon$.
\end{theorem}

\textit{Proof Outline}:
The main LT construction idea is visualized in Fig.~\ref{fig:lt-schematic}~(c) in the appendix. 
For every target neuron, multiple approximating copies are created in the respective layer of the LT to serve as basis for modified subset sum approximations (see Lemma~\ref{lemma:subset}) of the parameters that lead to the next layer.
In line with this approach, the first layer of the LT consists of univariate blocks that create multiple copies of the input neurons.
In addition to Lemma~\ref{lemma:subset}, also the total number of subset sum approximation problems $\rho$ that have to be solved needs to be re-assessed for ER source networks, as this influences the probability of LT existence.
This modification is driven by the same factor $1 / \log(1 / (1-p))$. 
The full proof is given in Appendix~\ref{proof-slts}.

With our SLT existence results, we have provided our first example of how to generally turn dense to sparse deep learning methods into sparse to sparse schemes.
Next, we also validate the idea to start pruning from a random ER mask in experiments.


%% file: expt.tex
\section{Experiments}
\label{experiments}

To verify our theoretical insights, we conduct experiments in standard settings on common benchmark data (CIFAR10, CIFAR100 \citep{cifar10-100} and Tiny ImageNet \citep{tiny-imagenet}) and neural network architectures (ResNet \citep{resnet} and VGG \citep{vgg}). 
Details on the setup can be found in App.~\ref{app:expt-setup} . 
We always report the mean over 3 independent runs. 
Due to space constraints, confidence intervals are reported in the appendix alongside additional experiments.

Our main objective is to showcase the expressiveness of ER networks with three kinds of experiments.
First, we highlight that a randomly pruned network with carefully chosen layerwise sparsity ratios are competetive and sometimes even outperform state-of-the-art pruning methods like Iterative Magnitude Pruning (IMP) \citep{weak-lth} (see Appendix \ref{app:dense-sparse-baseline})
Second, we verify that ER networks can serve as promising starting point of further sparsification by pruning within the initial ER network.
Third, we apply the same principle for strong lottery tickets (SLTs) and present the first sparse to sparse training results in this context.

\textbf{The Performance of Random Pruning}
To complement \citet{random-pruning-vita}, we conduct experiments in higher sparsity regimes $ \geq 0.9$ to test the limit up to which random ER networks are a viable alternative to more advanced but computationally expensive pruning algorithms. 
\citet{sanity-check-1,sanity-check-jackpot} have shown that randomizing the layerwise mask of pruned networks obtained with state-of-the-art pruning algorithms are often competitive and present strong baselines.
The corresponding sparsity ratios are computationally cumbersome to obtain and thus of reduced practical interest.
We still report comparisons with sparsity ratios obtained by randomized Snip \citep{snip}, Iterative Synflow \citep{synflow}, and IMP \citep{weak-lth} to demonstrate that the best performing sparsity ratios for ER masks are often different from the ones obtained from iteratively pruned tickets.
The previous state of the art is usually defined by ERK \citep{evci-rigl, granet}.
In addition, we propose two methods to choose layerwise sparsities, \textit{balanced} and \textit{pyramidal}, which often improve the performance of ER networks (see Table~\ref{table:results-vgg-cifar10} and further results for ResNets on CIFAR10 and 100 in App.~\ref{app:wlt-resnet-cifar}). 
Exemplary sparsity ratios are visualized in Fig.~\ref{fig:layerwise-c10}.
The pyramidal and balanced methods are competitive and even outperform ERK in our experiments for sparsities up to $0.99$. 
Importantly, they also outperform layerwise sparsity ratios obtained by the expensive iterative pruning algorithms Synflow and IMP. 
However, for extreme sparsities $1-p \geq 0.99$, the performance of ER networks drops significantly and even completely breaks down for methods like \textit{ER Snip} and \textit{pyramidal}.
We conjecture that \textit{ER Snip} and \textit{pyramidal} are susceptible to layer collapse in the higher layers and even flow repair (see App.~\ref{flow-preserve}) cannot dramatically increase the network's expressiveness.
The general limitations that we encounter at higher sparsities, however, are expected based on our theory.
These can be partially remedied by using the rewiring strategy of Dynamic Sparse Training (DST) \cite{evci-rigl}.

\begin{table}[h!]
\centering
\begin{tabular}{c |  c  c  c  c} 
Sparsity & $0.9$ & $0.99$ & $0.995$ & $0.999$ \\
\midrule 
Pyramidal & $92.9$ & $\bm{90.4}$ & $87.8$& $10$\\
Balanced & $\bm{93.2}$&  $89.3$& $\bm{85.9}$& $\bm{68.7}$\\
Uniform & $91.3$& $82.7$ &$73.7$  & $14.2$\\
ERK & $92.7$& $87.8$ & $84.5$ & $59.2$ \\
Snip (ER) & $\bm{93.2}$& $26.3$  & $10$& $10$\\ 
Synflow (ER) & $91.4$& $86.6$ & $84$ & $63.8$ \\ 
IMP (ER) & $90$& $90.2$ & $79$& $10$\\
\bottomrule
\end{tabular}%
\vspace{0.1cm}
\caption{\textit{ER networks with different layerwise sparsities on CIFAR10 with VGG16.} We compare test accuracies of our layerwise sparsity ratios \textit{balanced} and \textit{pyramidal} with uniform ones, ERK, and ER networks with layerwise sparsitiy ratios obtained by IMP, Iterative Synflow and Snip (denoted by ER). Confidence intervals are reported in Appendix \ref{app:wlt-resnet-cifar}.}
\label{table:results-vgg-cifar10}
\end{table}

\textbf{Dynamical Sparse Training} 
To improve randomly pruned networks at extremely high sparsities, we employ the RiGL algorithm \citep{evci-rigl} to obtain Table \ref{table:const_results}.
First, we only rewire edges, which allows us to start from relatively sparse networks.
Simply by redistributing edges, the performance of the ER network can be improved.
In particular, initial balanced or pyramidal sparsity ratios seem to be able to improve the performance of RiGL.
Table \ref{table:granet} in the appendix demonstrates that also starting RiGL (prune + rewire) from much higher sparsities of upto $0.9$ is possible without significant losses in accuracy, which highlights the utility of random ER masks even at extreme sparsities.

\begin{table}[h!]
\centering
\begin{tabular}{ l  |  l  c  l c  l c} 
Sparsity & \multicolumn{2}{c}{ $0.99$} &  \multicolumn{2}{c}{ $0.995$} & \multicolumn{2}{c}{ $0.999$} \\
\cmidrule{1-7}
  Rewired & $\times$ & \checkmark & $\times$ & \checkmark & $\times$ & \checkmark \\
 \midrule 
 ERK  & $87.8$ & $90.8$ & $84.5$& $88.3$& $59.2$ & $74.1$\\ 
 Balanced & $89.3$ & $91.4$& $85.9$ & $89.3$& $\mathbf{68.7}$ & $\mathbf{78.9}$\\ 
 Pyramidal & $\mathbf{90.4} $ & $\mathbf{92}$& $\mathbf{87.8}$ & $\mathbf{90.6}$& $10$& $9.8$\\ 
\end{tabular}%
\vspace{0.1cm}
\caption{\textit{ER networks rewired with DST:} Test Accuracies for an $ER(\mathbf{p})$ VGG16 with a fixed mask and after rewiring edges with RiGL \citep{evci-rigl,random-pruning-vita} on CIFAR10. Confidence intervals are reported in Appendix \ref{app:dst}.} 
\label{table:const_results}
\end{table}

\textbf{Sparse to Sparse Training with ER networks}
We verify that ER networks can serve as a promising starting point of further sparsification schemes that prune within the ER network as explained by Fig.~\ref{fig:er-explained}. 
Effectively, this idea can turn any dense-to-sparse training scheme into a sparse to sparse one.
As representative for an iterative pruning approach we study IMP and for continuous sparsification 
scheme we employ Soft Threshold Reparametrization (STR) \citep{kusupati2020soft}.
In Tables \ref{table:str-er} and \ref{table:imp-er}, we observe that we can start training with an ER mask of sparsity upto $0.9$ and prune the network further without much loss in performance.
For both STR and IMP,  pruning an ER network of sparsity 0.7 on CIFAR10 results in the same performance that we would obtain if we prune a dense network instead.
Our experiments show that for both STR and IMP, in particular, balanced initial pruning ratios can boost the performance of the general approach.

\begin{table}[h!]
\centering
\begin{tabular}{l | c c c c}
Initial Sparsity & $0.7$ &$0.7$ &$0.8$ &$0.9$  \\
\midrule
Final Sparsity& $0.96$ & $0.997$ & $0.997$ & $0.998$\\
\midrule
Balanced &$94.11$ &$\mathbf{90.7}$ & $\mathbf{90.28}$& $\mathbf{89.47}$\\
Pyramidal &$94.18$ &$90.1$ & $89.52$& $88.87$\\
ERK &$\mathbf{94.33}$ &$90.12$ & $89.51$&$88.25$\\
Uniform &$93.74$ &$88.92$ &$87.89$ &$86.07$\\
STR (ER) &$93.89$ &$89.31$ &$87.87$ &$85.86$\\

\end{tabular}
\caption{\textit{Sparse to sparse training with Soft Threshold Reparametrization in ER networks:} Results on a ResNet18 trained on CIFAR10.
STR (ER) denotes sparsity ratios obtained by STR.
For reference, starting from a dense network STR achieves $94.66\%$ and $90.95\%$ at sparsity $0.9$ and $0.993$ respectively.
See Appendix \ref{app:sparse-sparse} for confidence intervals.}
\label{table:str-er}
\end{table}

\begin{table}[h!]
\centering
\begin{tabular}{l | c c c c }
Initial Sparsity & $0.7$ &$0.7$ & $0.8$ & $0.9$\\
\midrule
Final Sparsity & $0.9$ & $0.99$ & $0.93$ &$0.97$ \\
\midrule
Balanced &$93.54$  &$90.72$ &$93.14$ &$91.89$ \\
Pyramidal &$\mathbf{93.65}$  &$\mathbf{92.23}$ &$93.24$ &$92.23$ \\
ERK &$93.5$  &$90.95$ &$\mathbf{93.57}$ &$\mathbf{93.21}$ \\
Uniform &$93.18$  &$90.15$ &$92.62$ &$90.41$ \\
 
\end{tabular}
\caption{\textit{Sparse to sparse training with Iterative Magnitude Pruning in ER networks:} Results on a ResNet18 trained on CIFAR10.
For reference, starting from a dense network IMP achieves $93.38\%$ and $91.39\%$ at sparsity $0.9$ and $0.99$ respectively.
See Appendix \ref{app:sparse-sparse} for confidence intervals.}
\label{table:imp-er}
\end{table}

\textbf{Experiments for SLTs} 
Similarly to our previous experiments, we can also prune a random ER mask to obtain SLTs. 
We use the edge-popup \cite{ramanujan-slth} algorithm to verify our theoretical derivations.
Table~\ref{table:slt_results} presents evidence for the fact that the search for SLTs does not need to be computationally as expensive as dense training.
Remarkably, we can start with a sparse ER network of up to $0.8$ sparsity instead of a dense one and still achieve competitive performance in finding a SLT with final sparsity $0.9$. 
Additional experiments are reported in the appendix (see Tables \ref{table:slt_vgg} and \ref{table:resnet110-slt}). 
\begin{table}[h!]
\centering
\begin{tabular}{ c | c  c c  c } 
Initial Sparsity &$0.7$ &$0.5$ &$0.8$ &$0.5$ \\
\midrule
Final Sparsity  &   $0.9$ &  $0.95$ &  $0.95$ &  $0.99$  \\
 \midrule
 Uniform &  $88$  & $87.8$   & $\mathbf{88.1}$  & $87.9$\\
 Balanced &$\mathbf{88.06}$ &$87.93$ &$87.86$ &$87.93$ \\
 Pyramidal &$87.73$ &$\mathbf{88.02}$ &$87.95$ &$\mathbf{87.97}$ \\
ERK &$88.04$ &$87.76$  &$88.02$ &$87.85$ \\
\end{tabular}
     \caption{\textit{ER networks for Strong Lottery Tickets}: Test accuracies of SLTs obtained by  edge-popup (EP) \citep{ramanujan-slth} pruning a sparse ER ResNet18 on CIFAR10. 
     Starting dense (see the App.~\ref{app:SLTexp}), EP achieves $87.86 \%$ accuracy for sparsity 0.9.} 
     \label{table:slt_results}
\end{table}

\textbf{Experiments on Diverse Tasks}
While most of our experiments are focused on image classification tasks, our theoretical insights are more general and apply to diverse target and source network structures. 
To demonstrate the broader scope of our results, we provide additional experiments for sparse to sparse training on ImageNet, Graph Convolutional Networks, algorithmic data and tabular data in Appendix \ref{app:expt-additional}.
Consistently, we find that pruning sparse random networks achieve competitive performance compared to pruning a dense network.

\section{Conclusions}
We have systematically explained the effectiveness of random pruning and thus provided a theoretical justification for the use of Erdös-Rényi (ER) masks as strong baselines for lottery ticket pruning and as starting point of dynamic sparse training. 
Our theory implies that random ER networks are as expressive as dense target networks if they are wider by a logarithmic factor in their inverse sparsity.
Our constructions suggest that random pruning, even though computationally cheap, does not achieve optimal sparsity but has great potential for further pruning.
This finding is also of practical interest, as initial sparse random sparse masks can avoid the computationally expensive process of pruning a dense network from scratch. 
As exemplary highlight, we have applied this insight to strong lottery tickets. 
We have proven theoretically and demonstrate experimentally that pruning for strong lottery tickets can be achieved by sparse to sparse training schemes. 

%% file: app.tex
\section{Appendix}

\label{appendix}

\subsection{Flow Preservation to prevent layer collapse in ER networks}
\label{flow-preserve}
Targeted pruning is known to be susceptible to layer collapse or just a sub-optimal use of resources (given in form of trainable parameters), when intermediary neurons receive no input despite nonzero output weights or zero output weights despite nonzero input weights.
To avoid this issue, \citep{synflow} has derived a specific data-independent pruning criterion, i.e., synaptic flow.
Yet, flow preservation can also be achieved with a simple and computationally efficient random repair strategy that applies to diverse masking methods, including random ER masking.

The main idea behind this algorithm is to connect neurons (or filters) with zero in- or out-degree with at least one other randomly chosen neuron (or filter) in the network.
To preserve the global sparsity, a new edge can replace a random previously chosen edge. 
Alternatively, ER networks with flow preservation could also be obtained by rejection sampling, which is equivalent to conditioning neurons on nonzero in- and out-degrees.
To still meet the target density $p_l$, the ER probability $\tilde{p}_l$ would need to be appropriately adjusted. 
Our experiments reveal, however, that most randomly masked standard ResNet and VGG architectures usually perserve flows with high probability for different layerwise sparsity ratios up to sparsities $\approx 0.95$ (see Appendix~\ref{appendix:flow-preserve}). 
The most problematic layers are the first and the last layer if the number of input channels and output neurons is relatively small.
In consequence, most pruning schemes keep these layers relatively dense in general.
In our theoretical derivations, we assume flow preservation in the first layer.

\label{appendix:flow-preserve}
\begin{figure*}[h!]
    \centering
    \begin{subfigure}[b]{\textwidth}
        \centering
        \includegraphics[width=0.5\textwidth]{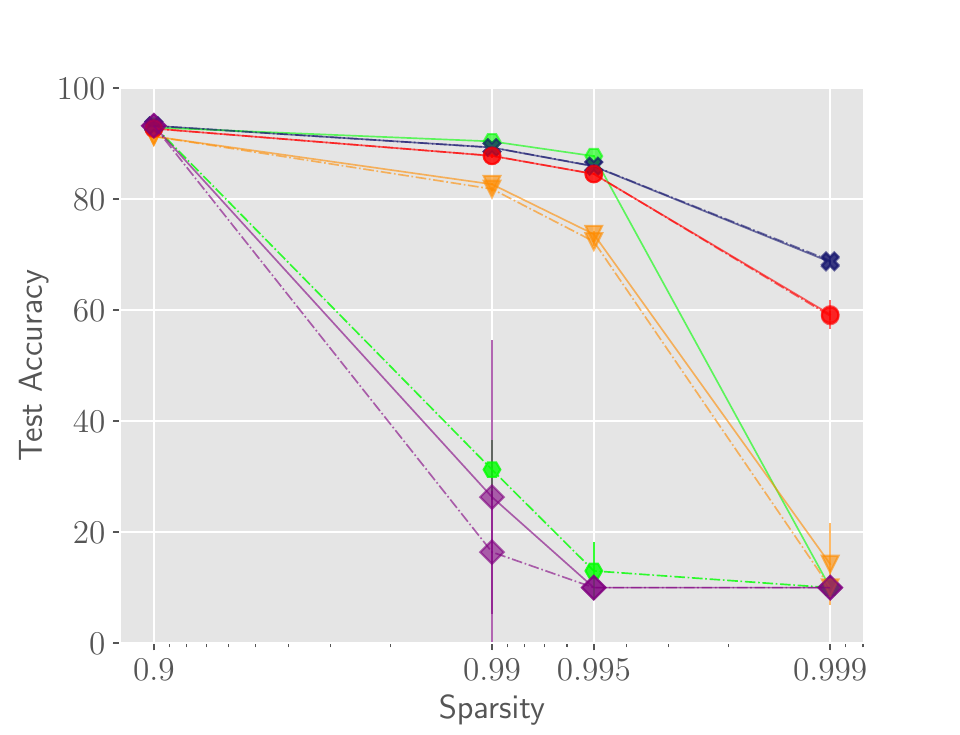}
    \end{subfigure}
    \includegraphics[width=\textwidth]{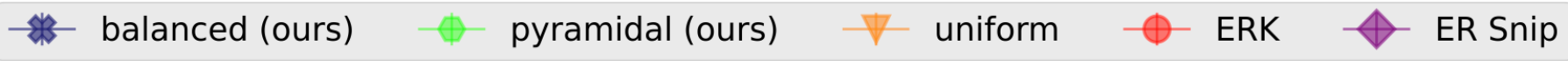}
    \caption{\textit{Flow Comparison}: We compare the results of ER networks for each layerwise sparsity method with and without flow preservation. Solid lines denote that flow is preserved while dotted lines show the corresponding method without flow preservation for a VGG16 on CIFAR10.}
    \label{fig:fig_flow}
\end{figure*}
We propose two methods to achieve flow preservation, which guarantees that every neuron (or filter) has at least in-degree and out-degree $1$. 

\textbf{Rejection Sampling}: We can resample the mask edges $s_{ij,ER}^{(l)}$ of the neurons (filters) that have a zero in-degree or a zero out-degree till there is at least one in-degree and one-out-degree for that neuron.

\textbf{Random Addition}: We randomly add an edge to a neuron with zero in-degree or out-degree. 
While this method adds an extra edge in the network, the total number of edges that need to be added are usually negligible in practice.

We verify the number of corrections required in an ER network to preserve flow. 
Notice that in most cases ER networks inherently preserve flow.
For each of the used layerwise sparsity ratios, we calculate the number of connections (edges) added in the network to ensure that every neuron (or filter) has at least in-degree and one out-degree $1$ using the Random Addition method. 
Tables \ref{tab:num_corrected_resnet} and \ref{tab:num_corrected_vgg} show the results. 

\begin{table}[H]
    \centering
    \begin{tabular}{c | c c c c c}
    \multirow{2}{*}{ResNet50}& \multicolumn{5}{c}{Sparsity} \\
    \cmidrule{2-6} 
     & $0.5$ & $0.8$& $0.9$& $0.99$& $0.999$ \\
    \midrule
      Uniform & $0$ & $0.33$& $1$& $51.67$& $108$ \\
      ERK & $0$ & $0$& $0$& $29.33$& $105.67$ \\
      ER Snip & $0$ & $0$& $10.67$& $59$& $87$ \\
      Balanced (ours) & $0$ & $0$& $0$& $23.33$& $92.33$ \\
      Pyramidal (ours) & $0$ & $0$& $1$& $47.67$& $91$ \\
    \end{tabular}
    \caption{Average number of mask edges added by flow correction in the ResNet18 network for CIFAR100 across three runs.}
    \label{tab:num_corrected_resnet}
\end{table}

\begin{table}[H]
    \centering
    \begin{tabular}{c | c c c c c}
    \multirow{2}{*}{VGG19} & \multicolumn{5}{c}{Sparsity} \\
    \cmidrule{2-6}
     & $0.5$ & $0.8$& $0.9$& $0.99$& $0.999$ \\
    \midrule
      Uniform & $0$ & $0.33$& $1.33$& $3$& $34$ \\
      ERK & $0$ & $0$& $0$& $0$& $30.33$ \\
      ER Snip & $0$ & $0$& $0$& $14.33$& $25$ \\
      Balanced (ours) & $0$ & $0$& $0$& $0$& $23.33$ \\
      Pyramidal (ours) & $0$ & $0$& $1$& $8$& $22$ \\
    \end{tabular}
    \caption{Average number of mask edges added by flow correction in the VGG19 network for CIFAR100 averaged across three runs. Note that the number of flow corrected neurons (filters) is negligible in comparison to the number of nonzero parameters in VGG19 even for the lowest density of $0.001$, which is $1,38,000$ parameters.}
    \label{tab:num_corrected_vgg}
\end{table}
Our analysis shows that \textit{flow preservation} is an important property that avoids layer collapse in sparse networks and is inherently satisfied in reasonable sparsity regimes $\approx 0.9$. 
It has a similar effect as making the final layer and the initial layer dense during pruning, which is followed in some pruning algorithms \citep{random-pruning-vita}.

Figure \ref{fig:fig_flow} compares the different layerwise sparsity methods for ER networks with and without flow preservation.
Our results show that flow preservation is especially important in Pyramidal and ER Snip methods.
Both these methods have a higher sparsity in the final layer which leads to performance problems in case of high global sparsities. 
Flow preservation is able to address this partially so that a clear improvement is visible for the pyramidal method at sparsities $0.99$ and $0.995$.

\subsection{Proof for Existence of Strong Lottery Tickets in ER networks}
\label{proof-slts}
\begin{figure*}[h!]
\centering
\includegraphics[width=0.9\textwidth]{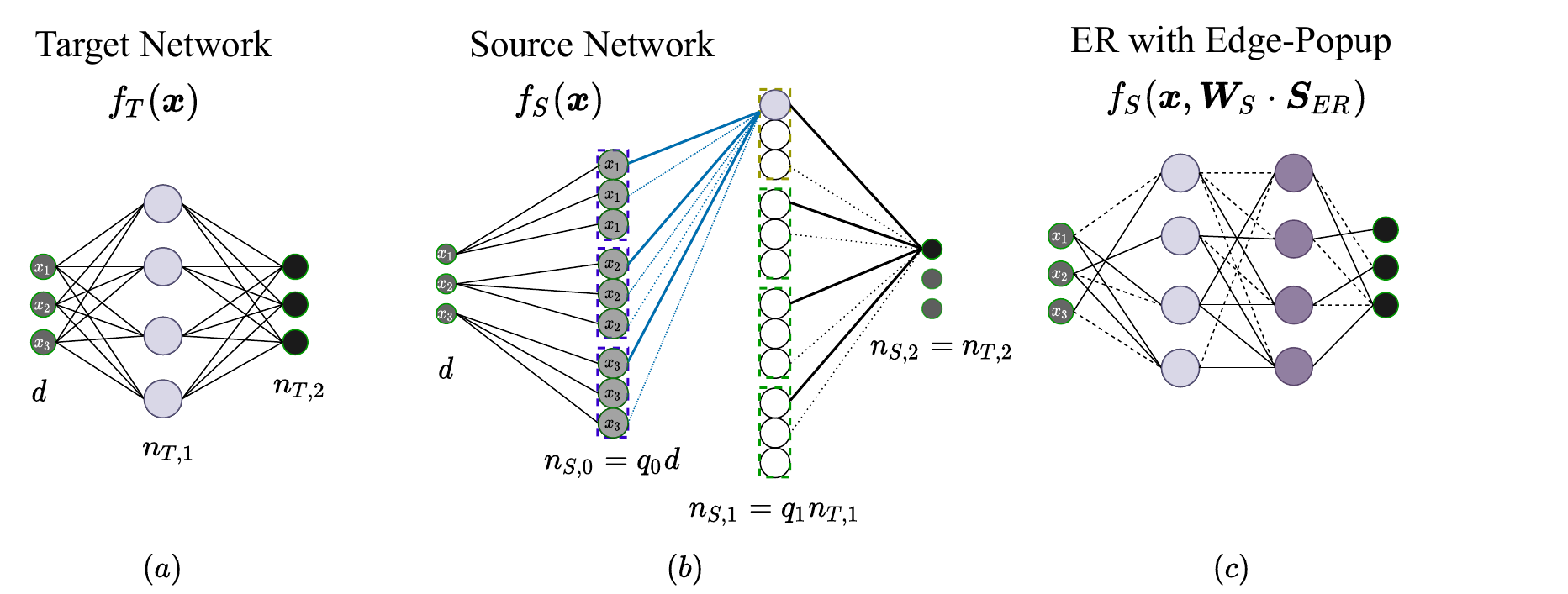}
\caption{\textit{Expressivity in ER networks:} In (a), $f_T(\bm{x})$ is a single layer target network. $(b)$ visualizes the source ER network $f_S(\bm{x})$, which contains a sparse network that represents the target. (c) shows a strong LT contained withing an ER network.
The figure shows connections for only one neuron in every layer of $f_S$ for simplicity. 
Both \textit{dotted} and \textit{solid} lines belong to the random mask $\bm{S}_{ER}$, while the \textit{solid} lines belong to nonzero weights of the final sparse network ($\bm{S_{P}}$). 
}
\label{fig:lt-schematic}
\end{figure*}

As discussed in the main manuscript, most SLT existence proofs that derive a logarithmic lower bound on the overparametrization factor of the source network utilize subset sum approximation \citep{lueker-subsetsum} in the explicit construction of a lottery ticket that approximates a target network \citep{pensia-subset-sum,uniExist,convexist,cnnexist,depthexist}.
We can transfer all of these proofs to ER source networks by modifying the subset sum approximation results to random variables that are set to zero with a Bernoulli probability $p$ to account for randomly missing links in the source network.
We just have to replace Lueker's subset sum approximation result by Lemma~\ref{lemma:subset} in the corresponding proofs.
For simplicity, we have formulated it for uniform random variables and target parameters $z \in [-1,1]$ but it could be easily extended to random variables that contain a uniform distribution (like normal distributions) and generally bounded targets as in Corollary 7 in \citep{uniExist}.
For convenience, we restate Lemma~\ref{lemma:subset} from the main manuscript:

\begin{lemma}[Subset sum approximation in ER Networks]
Let $X_1, ..., X_{n}$ be independent, uniformly distributed random variables so that $X_i \sim U([-1,1])$ and $M_1, ..., M_n$ be independent, Bernoulli distributed random variables so that $M_i \sim \text{Ber}(p)$ for a $p>0$.
Let $\epsilon, \delta \in (0,1)$ be given.
Then for any $z \in [-1,1]$ there exists a subset $I \subset [n]$ so that with probability at least $1-\delta$ we have $|z - \sum_{i \in I} M_i X_i| \leq \epsilon$ if
\begin{align}
    n \geq C \frac{1}{\log(1 / (1 - p))} \log\left(\frac{1}{ \min\left(\delta, \epsilon\right)} \right).
\end{align}
\end{lemma}
\textbf{Proof}
Random variables $\tilde{X}_{i} = M_i X_i$ do not contribute to the approximation of a target value $z$, if they are zero and thus in particular in the case that $M_i = 0$, which happens with probability $1-p$ for each index $i$. 
We can thus remove all the variables $\tilde{X}_{i}$, for which $M_i = 0$. 
After a change of indexing, we arrive at a subset $\tilde{X}_{1}, ..., \tilde{X}_{K}$ of $K$ random variables, which are uniformly distributed as $\tilde{X}_{i} = M_i X_i = X_i \sim U([-1,1])$, since $M_i$ is independent of $X_i$.
The number of variables $K$ follows a binomial distribution, $K \sim \text{Bin}(n,p)$, since $M_1, ..., M_n$ are independent Bernoulli distributed.

For fixed $K=k$, \citet{lueker-subsetsum} has proven that there exists constants $a_k >0$ and $b_k >0$ so that the probability that the approximation is not possible is of the form $\mathbb{P}\left((\forall I \subset [k]) \; |z-\sum_{i\in I} \tilde{X}_i| > \epsilon'  \right) \leq a_k \exp(-b_k k)/\epsilon'$.

Using this result and defining $a := \max_{k \in [n]} a_k > 0$ and $b := \min_{k \in [n]} b_k > 0$, we just have to take an average with respect to the random variable $K \sim \text{Bin}(B,p)$. 
\begin{align*}
    \mathbb{P}\left((\forall I \subset [n]) \; |z-\sum_{i\in I} \tilde{X}_i| > \epsilon'  \right) & \leq \sum^n_{k=0} \frac{a_k}{\epsilon'}  \exp(-b_k k) {n \choose k} p^k (1-p)^{n-k} \\
    & \leq  \frac{a}{\epsilon'}  \sum^n_{k=0} {n \choose k} \exp(-b k) p^k (1-p)^{n-k}\\
    & =  \frac{a}{\epsilon'} [1-p(1 - \exp(-b))]^n
\end{align*}
To ensure the subset sum approximation is feasible with probability of at least $1-\delta'$ we need to fulfill 
\begin{align*}
    \frac{a}{\epsilon'}[1 - p(1 - \exp(-b))]^n \leq  \delta'. 
\end{align*}
Solving for $n$ leads to
\begin{align*}
    n \geq \frac{1}{\log\left(\frac{1}{1 - p(1 - \exp(-n))}\right)}\log\left(\frac{a}{ \delta'\epsilon'} \right).
\end{align*}
This inequality is satisfied if 
\begin{align*}
    n \geq  C \frac{1}{\log(1 / (1 - p))} \log\left(\frac{1}{ \min\{\delta',\epsilon'\} }\right)
\end{align*}
for a generic constant $C > 0$ that depends on $a$ and $b$.

With this modified subset sum approximation, we show next that in comparison with a complete source network, an ER network needs to be wider by a factor $\frac{1}{\log(1 / (1 - p))}$.
To provide an example of how to transfer an SLT existence proof, we focus on the construction by \cite{depthexist}.

Note that in all our theorems we assume that flow is preserved in the first layer, as it is reasonable to apply a simple and computationally cheap flow preservation algorithm after drawing a random mask (see Appendix~\ref{appendix:flow-preserve}).
This algorithm just ensures that all neurons are connected to the main network and are thus useful for training a neural network.

If we do not assume that flow is preserved, some neurons in the first layer might be disconnected from all input neurons with probability $(1-p_0)^d$. 
Disconnected neurons could simply be ignored in the LT construction. 
Their share is usually negligible but, technically, without flow preservation, we would need to ensure that $n_{S,1} \geq C (1-p_0)^d + n^*_{S,1}$, where $n^*_{S,1}$ denotes the bound on the width that we are actually going to derive.

\begin{theorem}[Existence of SLTs in ER Networks]
Let $\epsilon, \delta \in (0,1)$, a target network $f_T$ of depth $L$, an $\text{ER}(\mathbf{p})$ source network $f_S$ of depth $L+1$ with edge probabilities $p_l$ in each layer $l$ and iid initial parameters $\bm{\theta}$ with $w_{ij}^{(l)} \sim U([-1,1]), b_i^{(l)} \sim U([-1,1])$ be given. 
Then with probability at least $1- \delta$, there exists a mask $\bm{S}_{P}$ so that each target output component $i$ is approximated as $\max_{\bm{x}\in \mathcal{D}}\|f_{T,i}(\bm{x}) - f_{S,i}(\bm{x}; \bm{W}_{S}\cdot\bm{S}_{P} )\| \leq \epsilon $ if 
\begin{align*}
    n_{S,l} \geq C\frac{n_{T,l}}{\log\left(1/(1-p_{l+1})\right)}\log \left(\frac{1}{\min\{\epsilon_{l}, \delta/\rho\}}\right)
\end{align*}
for $l \geq 1$, where $\epsilon_{l} = g(\epsilon, f_T)$ is defined in Equation~(\ref{eq:eps}) and $\rho = \frac{CN_{T}^{1+\gamma}}{\log(1 / (1-\min_l p_l))^{1+\gamma}} \log(1 / \min\{\min_l\epsilon_l, \delta\})$ for any $\gamma \geq 0$. We also require $n_{S,0} \geq C d \frac{1}{\log(1/(1-p_1))}\log\left(\frac{1}{\min\{\epsilon_1, \delta / \rho\}}\right)$, where $C>0$ denotes a generic constant that is independent of $n_{T,l}$, $L$, $p_l$, $\delta$, and $\epsilon$.
\end{theorem}

Here, $\epsilon_l = g(\epsilon)$ is defined in accordance with Lemma 5.1 in \citep{depthexist}:
\begin{align}\label{eq:eps}
    \epsilon_l = g(\epsilon, f_T) = 
    \frac{\epsilon}{n_{T,L}L}\left[(1 + B_{l-1})(1 + \frac{\epsilon}{L})
    \prod_{k=l+1}^{L-1}(||W_T^{(k)}||_{\infty} + \frac{\epsilon}{L})\right]^{-1}, \ 
    B_l := \sup_{x \in \mathcal{D}} ||\bm{x}_T^{(l)}||_1.
\end{align}

\textbf{Proof} To prove the existence of strong lottery tickets in ER networks, we modify the proof by \citep{depthexist} for complete fully-connected networks.
We first answer the question, how the fact that random weights are set irreversibly to zero, changes our construction.
Fig. \ref{fig:lt-schematic} visualizes the general schematic.
The general idea is that we have to create multiple copies $\rho_l$ of each target neuron in the LT, as these will enable the approximation of target parameters by utilizing subset sum approximation as modified by Lemma~\ref{lemma:subset}. 
First, as Fig. \ref{fig:lt-schematic} visualizes, we have to argue why and how we can create univariate blocks in the first layer or in general $2L$ constructions.
In this case, a target layer is approximated by two appropriately pruned layers of the source network.
The first of these two source layers contains only univariate neurons that form blocks that consist of neurons of the same type, which correspond to the same input target neuron $i$. 
All weights that start in the same block $i$ and end in the same neuron $j$ can then be utitilzed to approximate the target parameter $w_{T,ji}$. 
The required univariate blocks can be easily realized by pruning if flow is preserved. 
The reason is that each neuron in source layer $l=0$ has at least one in-coming edge, which can survive the pruning.
Since this edge could be adjacent to any of the input neurons with the same probability, we can always find enough neurons in Layer $l=0$ that point to any of the input neurons and this allows us to form univariate blocks of similar size $B$.

Second, we have to analyze how the construction of each following target layer is affected by randomly missing edges in the source network.
Each target weight $w^{(l)}_{T,ij}$ can be approximated by $w^{(l)}_{T,ij} \approx \sum_{j'\in I} m^{(l)}_{S,i'j'} w^{(l)}_{S,i'j'}$, where the neuron $i'$ in the LT approximates the target neuron $i$ and the neuron $j'$ in the LT approximates the target neuron $j$.
The subset $I$ is chosen based on a modified subset sum approximation and informs the mask of the LT.
Thus, $I$ exists according to Lemma~\ref{lemma:subset}, since the initially random mask entries of the source network $m^{(l)}_{S,i'j'}$ are Bernoulli distributed with probability $p_l$.

The second issue that needs to be modified for ER networks is the analysis of the number of required subset sum approximation problems $\rho$.
As explained before, the main idea of the construction is to create $\rho_l$ copies of each target neuron in target Layer $l$ in Layer $l$ of the LT.
These copies serve then multiple subset sum approximations to approximate the target neurons in the next layer in a similar way as the univariate blocks of the first layer. 
This, however, increases the total number of subset sum approximation problems $\rho$ that need to be solved and that influence the probability with which we can solve all of them. 
Using a union bound, we can spend $\delta / \rho$ on every approximation with a modified $\rho$ for ER networks.
Similar to \citep{depthexist}, we can derive a lower bound on $\rho_l$ in the subsequent layers, so that the subset sum approximation is feasible for every parameter of layer $l$ when the block size $B$ is
\begin{align*}
    B \geq \frac{1}{\log(1 / (1 - p_l))} \log \left(\frac{a}{ \frac{\delta'}{\rho}\epsilon'}\right)
\end{align*}
so that with an appropriately chosen constant $C$ we have
\begin{align*}
    B \geq \frac{C}{\log(1 / (1 - p_l))} \log \left(\frac{1}{\min\{ \frac{\delta'}{\rho},\epsilon'\}}\right)
\end{align*}
so that it follows in total that 
\begin{align*}
    n_{S,l} \geq C \frac{n_{T,l}}{\log(1 / (1 - p_{l+1}))}\log\left(\frac{1}{\min\{\epsilon_l. \delta/\rho\}}\right)
\end{align*}

The remaining objective is to find a $\rho \geq \rho' = \sum_{l=1}^{L} \rho'_l$, where $\rho'$ is the factor of increased subset sum approximation problems required to approximate $L$ target layers with an ER source network and $\rho_l$ counts the number of parameters in each LT layer.

Following \citet{depthexist}'s method to identify $\rho$, we start with the last layer. 
The number $\rho_L$ of subset sum approximation problems that have to be solved to approximate the last layer determines the number of neurons required in the previous layer.
This in turn determines the required number of neurons in the layer before it, etc.
The last layer requires to solve exactly $\rho'_L = n_{T,L}n_{T,L-1}$ subset sum problems which can be solved with sufficiently high probability if $n_{S,L-1} \geq \frac{Cn_{T,L-1}}{\log(1 / (1 - p_{L}))} \log(1 / \min\{\epsilon_L, \delta / \rho'\})$.
As we would need maximally $\frac{C}{\log(1 / (1 - p_L))}\log(1 / \min\{\epsilon_L, \delta / \rho'\})$ sets of the target parameters in the last layer, we can bound $\rho'_{L-1} \leq \frac{CN_{L-1}}{\log(1 / (1 - p_L))}\log(1 / \min\{\epsilon_L, \delta / \rho'\})$.
Repeating the same argument for every layer, we derive 
$\rho'_{l} \leq \frac{CN_{l}}{\log(1 / (1 - p_{l+1}))}\log(1 / \min\{\epsilon_{l+1}, \delta / \rho'\})$.
In total, we find that $\rho' = \sum_{l=1}^L \rho'_l \leq \sum_{l=1}^{L}\frac{CN_l}{\log(1 / (1 - p_{l+1}))}\log (1 / \min\{\epsilon_{l+1}, \delta/\rho'\}) \leq \frac{CN_t}{\log(1 / (1 - \min_l p_{l}))}\log (1 / \min\{\min_l \epsilon_{l}, \delta/\rho\})$.
Here, $N_l = n_{T,L}n_{T,L-1}$ and $N_t = \sum_l N_l$.
A $\rho$ that fulfills $\rho \geq \frac{CN_t}{\log(1 / (1 - \min_l p_{l}))}\log (1 / \min\{\epsilon_{l+1}, \delta/\rho\}) $ will be sufficient.
It is easy to see that $\rho = \frac{CN_{T}^{1+\gamma}}{\log(1 / (1-\min_l p_l))^{1+\gamma}} \log(1 / \min\{\min_l\epsilon_l, \delta\})$ for any $\gamma \geq 0$ fulfills our requirement.

We have thus shown the existence of SLTs in ER networks following similar ideas as the proof of Theorem 5.2 by \citet{depthexist}.
Thus, our construction would also apply to more general activation functions than ReLUs.
Note that we could also follow the proof strategy of \citet{pensia-subset-sum} to show the existence of strong lottery tickets in ER networks. 
The key difference between the proofs of \citet{depthexist} and \citet{pensia-subset-sum} is how the subset sum base is created to approximate a target parameter.
\citet{pensia-subset-sum} use two layers for every layer in the target and create a basis set to approximate every target weight while \citet{depthexist} go one step further and create multiple subset sum approximations of every target weight to avoid the two layer construction.
In both these cases, the underlying subset sum approximation can be modified as shown above for ER networks and the same proof strategy as \citep{depthexist} or \citep{pensia-subset-sum} can be followed.
Similarly, we could also extend our proofs to convolutional and residual architectures \citep{convexist}.

\subsection{Representing a Single Hidden Layer Target Network with a Two Layer ER network}
\label{proof-single}

\begin{theorem}[Single Hidden Layer Target Construction]
Assume that a single hidden-layer fully-connected target network $f_T(\bm{x}) = \bm{W}_T^{(2)}\phi(\bm{W}_T^{(1)}\bm{x} + \bm{b}_T^{(1)}) + \bm{b}_T^{(2)}$, an allowed failure probability $\delta \in (0,1)$, source densities $\mathbf{p}$ and a $2$-layer ER source network $f_{S} \in \text{ER}(\mathbf{p})$ with widths $n_{S,0} = q_0d, n_{S,1} = q_1n_{T,1}, n_{S,2} = q_2n_{T,2}$ are given. 
If
\begin{align*}
   q_{0} \geq \frac{1}{\log(1 / (1-p_1))}\log\left(\frac{2 m_{T,1}q_1}{\delta}\right),  \\
   q_{1} \geq \frac{1}{\log(1 / (1-p_2))}\log\left(\frac{2 m_{T,2}}{\delta}\right) \text{ and } q_{2} = 1
\end{align*}
then with probability $1-\delta$, the random source network $f_S$ contains a subnetwork $\bm{S}_{P}$ such that $f_S(\bm{x},\bm{W}\cdot\bm{S}_{P}) = f_T$.
\end{theorem}

\textbf{Proof of Theorem \ref{thm:existence-single}}
A two hidden layer network can approximate a single hidden layer target network as explained in Section \ref{thm:existence-single}. 
$(q_0, q_1, q_2)$ are the overparametrization factors in each layer in the source network which ensure that we can find the links that we need in the ER network. 
Why would we need any form of overparametrization?
Different from the SLT construction, we do not need to employ multiple parameters to approximate a single parameter and thus do not use any subset sum approximation.
We choose the weights in the ER network such that they are exactly the corresponding weights of the target network.
Yet, we still need to prove that we can find all required nonzero entries in our mask.
To increase the probability that a target link exists, we also create multiple copies of input neurons.
As in the SLT construction, we prune the neurons in first layer to univariate neurons and choose the bias large enough so that the ReLU acts essentially as an identity function. 
$p_0 > 0$ can thus be arbitrary, as long as flow is preserved.
Note that $q_2 = 1$, as the output neurons for the source and target should be identical $n_{T, 2} = n_{S, 2}$. 
The last layer (output layer) in the target contain $n_{T, 2}$ neurons and the penultimate layer $n_{T, 1}$. 
In the source network, we create $q_1$ copies of each neuron in the second layer of the target network such that $n_{S, 1} = q_1 \times n_{T, 1}$. 
Our goal is to bound the width of Layer 1 in the ER network such that there is at least one nonzero edge in the ER network for every nonzero target weight.
To lower bound $q_1$, each nonzero weight $w_{T, ij}^{(2)}$ must have at least one nonzero weight (edge) in the source network with sufficiently high probability, i.e., every neuron in the output layer $n_{S,2}$ must have a nonzero edge to every block in the previous layer $n_{S,1}$ as explained in Figure \ref{fig:lt-schematic}.
The probability that at least one such edge exists for each output neuron is given as $\left(1- (1-p_2)^{q_1} \right)^{m_{T,2}} $.

Similarly, we can compute the probability that each neuron in the second layer of the source $n_{S, 1}$ has at least one nonzero edge to to each of the univariate blocks in the first layer as $\left(1- (1-p_1)^{q_0} \right)^{m_{T,1} \times q_1}$. 
Since each layer construction is independent from the other, the above probabilities can be multiplied to obtain the probability that we can represent the entire target network as 
\begin{align*}
    \prod^2_{l=0} \left(1- (1-p_l)^{q_{l-1}} \right)^{m_{T,l} q_{l}} \geq  1- \delta
\end{align*}

One way to fulfill the above inequality is to split the error between the two product terms,
\begin{align*}
    \left(1- (1-p_1)^{q_{0}} \right)^{m_{T,1} q_{1}} \geq  \left(1- \delta \right)^{\frac{1}{2}} \text{ and } \left(1- (1-p_2)^{q_{1}} \right)^{m_{T,2} q_{2}} \geq  \left(1- \delta \right)^{\frac{1}{2}}
\end{align*}

Both equations above are satisfied with $ 1- (1-p_2)^{q_{1}} \geq  \left(1- \frac{\delta}{2m_{T,2} q_{2}} \right)$ and $1- (1-p_1)^{q_{0}} \geq  \left(1- \frac{\delta}{2m_{T,1} q_{1}} \right)$. 
We can now solve for $q_i, i \in \{0,1\}$
\begin{align*}
    q_0 \geq \frac{1}{\log(1 / (1 - p_1))} \log\left(\frac{2m_{T,1}q_1}{\delta}\right)
\end{align*}
and 
\begin{align*}
    q_1 \geq \frac{1}{\log(1 / (1 - p_2))} \log\left(\frac{2m_{T,2}}{\delta}\right), \text{ since } q_2 = 1
\end{align*}

After having identified a representative link in the source ER network for each target weight, we next define the weights and biases for the source ER network.
Each representative link in the ER source network is assigned the weight of its corresponding target.
For the first layer in the source network, which is an univariate construction of the input, the weights are defined as $w_{S, ij}^{(0)} = 1$ and the bias is large enough so that all relevant inputs pass through the ReLU activation function as if it was the identity:
\begin{align*}
    w_{S, ij}^{(0)} = 1 \ \forall j \in \{1, 2, .., d\} \text{ and } i \in \{1, 2, .., n_{S,0}\},
\end{align*}
\begin{align*}
    b_{S,i}^{(0)} = 
    \left\{
    	\begin{array}{ll}
    		-a_1  & \mbox{if } a_1 \leq 0 \\
    		0 & \mbox{if } a_1 > 0
    	\end{array}
    \right.
    \text{for every } i \in \{1, 2, .., n_{S,0}\}.
\end{align*}
Recall that $a_1$ is defined as the lower bound of each input input component $\bm{x}$.
We compensate for this additional bias in the last layer.
Now for the second layer, every weight $w_{T, ij}^{(1)}$ in the target network is assigned to one of the nonzero mask entries in the ER source network that lead to the corresponding input block $j$ and output block $i$.
The remaining extra weights in the source are set to zero.
\begin{align*}
    w_{S, i' j'}^{(1)} = w_{T, ij}^{(1)}, i' \in \{q_{1}i, q_{1}i + 1, .., q_{1}i + q_{1}\} \text{ and } j' \in \{q_{0}j, q_{0}j + 1, .., q_{0}j + q_{0}\}
\end{align*}
for one pair of $i', j'$. 
The remaining connections between $i'$ and block $j$ can be pruned away, i.e., masked or set to zero. 
The bias of the second layer can be chosen so that it compensates for the extra bias added in the univariate construction of the first layer:
\begin{align*}
    \forall i' \in \{1, ..., n_{S,1}\} \  b_{S,i'}^{(1)} = b_{T,i}^{(1)} -  w_{T, ij}^{(1)} b_{S,j'}^{(0)}.  
\end{align*}

\subsection{Representing a target network of depth $L$ with ER networks}
\label{proof-general}


Extending our insight from the $2$-layer construction of the source network in the previous section, we provide a general result for a target network $f_T$ of depth $L$ and ER source networks with different layerwise sparsity ratios $p_l$.
While we could approximate each target layer separately with two ER source layers, we instead present a construction that requires only one additional layer so that $L_s = L+1$. 
This is in line with the approach used by \citet{depthexist,convexist} for SLTs.
But we have to solve two extra challenges.
(a) We need to ensure that a sufficient number of neurons are connected to the previous layer with nonzero edges.
(b) We have to show that the required number of potential matches for target neurons $q_l$ does not explode for an increasing number of layers. In fact it only scales logarithmically in the relevant variables.
\begin{theorem}[ER networks can represent $L$-layer target networks.]
Given a fully-connected target network $f_T$ of depth $L$, $\delta \in (0,1)$, source densities $\mathbf{p}$ and a $L+1$-layer ER source network $f_{S} \in \text{ER}(\mathbf{p})$ with widths $n_{S,0} = q_0d$ and $n_{S,l} = q_l n_{T,l}, l \in \{1, 2, .., L\}$, where
\begin{align*}
       q_{l} \geq \frac{1}{\log(1 / (1-p_{l+1}))}\log\left(\frac{L m_{T,l+1} q_{l+1} }{\delta}\right) \\
       \text{for } l \in \{0, 1, .., L-1\} \text{ and } q_L = 1,
    \end{align*}
then with probability $1-\delta$ the random source network $f_S$ contains a subnetwork $\bm{S}_{P}$ such that $f_S(\bm{x},\bm{W}\cdot\bm{S}_{P}) = f_T$.
\end{theorem}

\textbf{Proof:} Again we follow the same procedure of finding the smallest width for every layer in the source network such that there is at least one connecting edge between a target neuron copy and one of the copies in the previous layer. 
Repeating this argument for every layer starting from the output layer in reverse order gives us the lower bound on the factor $q_l$ in every layer $l \in \{0, 1, .., L\}$. 
We choose the weights of the sparse ER network such that for every target parameter there is at least one nonzero (unmasked) parameter in the source which exactly learns the required value. 

We now construct a source network $f_S(\bm{x})$ that contains a random subnetwork which replicates $f_T(\bm{x})$ with probability $1-\delta$. 
As explained in Section~\ref{proof-single}, we first construct an univariate layer (with index $l=0$) in the source network assuming flow preservation. 

Next, we calculate the overparametrization factor required for every layer in the source network using the same argument as \ref{proof-single} starting from the last layer and working our way backwards. 
The last output layer has the same number of neurons in both the source and the target, $n_{S,L} = n_{T,L}$. Hence, the required width overparametrization factor is $q_{L}=1$. 
In every intermediary layer, we create blocks of neurons that consist of $q_l$ replicates of the same target neuron.
How large should $q_l$ be?
In the second to last layer, the probability that each neuron in the output layer has at least one edge to each of the $q_{L-1}$ blocks in Layer $L-1$ is $(1 - (1 - p_L)^{q_{L-1}})^{m_{T,L}q_{L}}$. 
We can similarly compute this probability for every layer all the way to the input in the source network which ensures that there is at least one edge between a neuron in every layer and each of the $q_{l-1}$ blocks in the previous layer. 
The probability that Layer $l$ can be constructed is thus $(1 - (1 - p_l)^{q_{l-1}})^{m_{T,l}q_{l}}$. 
These events are independent and should hold simultaneously with probability $1-\delta$. 
The following inequality formalizes our argument
\begin{align*}
    \prod^L_{l=1} \left(1- (1-p_l)^{q_{l-1}} \right)^{m_{t,l} q_{l}} \geq  1- \delta
\end{align*}
One way to fulfill the above equation would be to ensure that
\begin{align*}
  \left(1- (1-p_l)^{q_{l-1}} \right)^{m_{T,l} q_{l}} \geq  \left(1- \delta \right)^{1/L} 
\end{align*}
for each layer and thus

\begin{align*}
  \left(1- (1-p_l)^{q_{l-1}} \right) \geq  \left(1- \delta \right)^{1/(m_{T,l}q_{l}L)} 
\end{align*}

This inequality is fulfilled if
\begin{align*}
    1- (1-p_{l})^{q_{l-1}} \geq  1- \frac{\delta}{m_{T,l}q_{l}L} 
\end{align*}

Solving for $q_{l-1}$ leads to
\begin{align*}
    q_{l-1} \geq \frac{\log\left(\frac{\delta}{m_{T,l}q_{l}L}\right)}{\log(1 - p_l)} = \frac{1}{\log (1 / (1 - p_l))} \log \left( \frac{Lm_{T,l}q_l}{\delta}\right).
\end{align*}
We can thus compute the required width overparametrization for every layer starting from the last one, where we know $q_L = 1$. 
Note, that $q_{l}$ depends on the logarithm of $q_{l+1}$ of the next layer, which ensures that $q_{l}$ does not blow up as depth increases. 

After making sure that the required edges exist in the ER network to represent every target weight, we still have to derive concrete parameter choices. 
It follows then that these choices of weights can be chosen, assuming they are a result of training the network.
Similar to the single hidden layer case, each representative link in the ER source network is assigned the weight of its corresponding target and the weights in the first univariate layer are set to 1. 
The biases in the univariate layer are chosen so that all the inputs pass through the ReLU activation.
The biases in the next layer compensate for the additional biases in the first layer.
\begin{align*}
    w_{S, ij}^{(0)} = 1 \text{ } \forall j \in \{1, 2, .., d \} \text{ and } i \in \{1, 2, ..., n_{S,0}\},
\end{align*}
\begin{align*}
    b_{S,i}^{(0)} = 
    \left\{
    	\begin{array}{ll}
    		-a_1  & \mbox{if } a_1 \leq 0, \\
    		0 & \mbox{if } a_1 > 0
    	\end{array}
     \right. 
    \text{for every } i \in \{1, 2, ..., n_{S,0} \}.
\end{align*}
For the subsequent layers in the source network $l \in \{1, 2, .., L\}$ the weights are  $w_{S, i' j'}^{(l)} = w_{T, ij}^{(l)}$, where $j'$ is randomly chosen among all the non-masked connections of $i'$ to block $j$ and
 $j' \in \{q_{l-1}j, q_{l-1}j + 1, .., q_{l-1}j + q_{l-1}\} \text{ and } i' \in \{q_{l}i, q_{l}i + 1, .., q_{l}i + q_{l}\}$.
The remaining connections between block $j$ and $i'$ can be pruned away or the weight parameters set to zero.
The biases are set to the corresponding target bias for layers $ l \in \{2, .., L\}$
\begin{align*}
     \forall i' \in \{q_{l}i, q_{l}i + 1, .., q_{l}i + q_{l}\}  \ \ \ b_{S,i'}^{(l)} = b_{T,i}^{(l)}
\end{align*}
but the second layer $l=1$ has an additional term to compensate for the bias in the first (univariate) layer:
\begin{align*}
    \forall i' \in \{q_{l}i, q_{l}i + 1, .., q_{l}i + q_{l}\} \ \ \ \ 
    b_{S,i'}^{(1)} = b_{T,i}^{(1)} - w_{T,ij}^{(1)}b_{S,j}^{(0)}. 
\end{align*}

\subsection{ER networks for Convolutional Layers}
\label{app:proof-conv}
We can also extend our analysis to ER networks with convolutional layers, where the number of channels need to be overparameterized by a factor of $1 / \log(1 / \text{sparsity})$. 

\begin{theorem}[ER networks can represent $L$ layer convolutional target networks]
\label{thm:existence-conv}
Given a target network $f_T$ of depth $L$ with convolutional layers $\bm{h}_{T, i}^{(l)} = \sum_{j=1}^{c_{l-1}}\bm{W}_{T, ij}^{(l)} * \bm{x}_{ij}^{(l-1)} + b_{T, i}^{(l)}, \bm{W}_{T} \in \mathbb{R}^{c_l \times c_{l-1}\times k_l}$, $\delta \in (0,1)$, a source density $\mathbf{p}$ and a $L+1$-layer ER source network $f_{S} \in \text{ER}(\mathbf{p})$ with convolutional layers $\bm{h}_{S, i}^{(l)} = \sum_{j=1}^{c_{l-1}}\bm{W}_{S, ij}^{(l)} * \bm{x}_{ij}^{(l-1)} + b_{S, i}^{(l)}, \bm{W}_{S} \in \mathbb{R}^{q_lc_l \times c_{l-1}\times k_l}$ where
\begin{align*}
       q_{l} \geq \frac{1}{\log(1 / (1-p_{l+1}))}\log\left(\frac{L m_{T,l+1} q_{l+1}}{\delta}\right) \\
       \text{for } l \in \{0, 1, .., L-1\} \text{ and } q_L = 1,
\end{align*}
then with probability $1-\delta$ the random source network $f_S$ contains a subnetwork $\bm{S}_{P}$ such that $f_S(\bm{x},\bm{W}\cdot\bm{S}_{P}) = f_T$.
\end{theorem}

\textbf{Proof}: 
 Similarly as in case of fully-connected ER source networks, we create $q_l$ copies of every output channel of the target $c_l$ in the source.
 Each channel copy in the source is assigned the weight of the corresponding target channel.
 Note that any tensor entry that leads to the same block is sufficient, since the convolution is a bi-linear operation so that $\sum_{i'\in I_i}\bm{W}_{i'j} * \bm{x}_i = \left(\sum_{i' \in I_i}\bm{W}_{i'j}\right) * \bm{x_i}$. 
Specifically, $\sum_{i' \in I_i} \bm{W}^{(l)}_{S,i'j}$ can represent a target element $w_{T,ije}^{(l)}$ if at least one weight $w_{S,i'je}^{(l)}$ is nonzero.

\begin{figure*}[h!]
    \centering

        \includegraphics[width=0.8\textwidth]{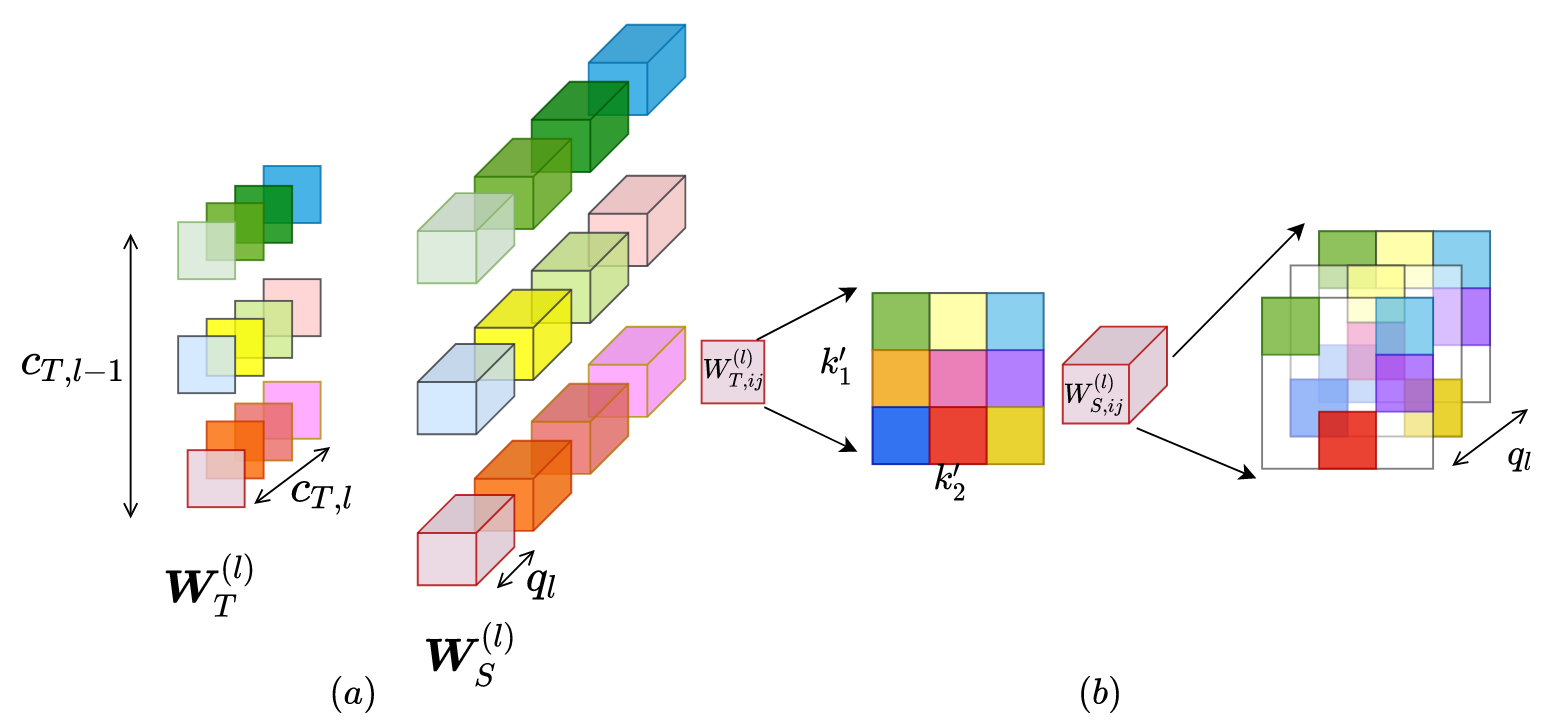}

    \caption{\textit{Construction of a convolutional target in an ER network}: For every output channel $c_{T,l}$ in the target convolutional weight tensor $\bm{W}_{T}^{(l)}$, we create $q_l$ copies in the source weight tensor $\bm{W}_{S}^{(l)}$ as shown on the left $(a)$. The width overparametrization is further elucidated in $(b)$ where each filter element of a target output filter has $q_l$ independent copies in the source, at least one of which is nonzero (unmasked). Coloured squares in $(b)$ show the nonzero parameters in the source ER network.}
    \label{fig:conv}
        
\end{figure*}

The linearity of convolutions allows us to construct a target filter by combining elements that are scattered between different input channels in the ER source network as shown in Figure \ref{fig:conv}. 
Using the same argument as the fully connected layer case, we bound the probability that at least one of the $q_l$ channels of every filter element in a convolutional weight tensor has a non-masked entry to a channel in the next layer.
As for fully-connected networks, we can create blocks of channels that correspond to replicates of the same target channel.
The first layer can be pruned down to univariate convolutional filters.
The probability that each layer can thus be reconstructed in the convolutional network can be bounded as:
\begin{align*}
  \left(1- (1-p_{l})^{q_{l-1}}\right)^{m_{T,l}q_{l}} \geq  \left(1- \delta\right)^{1/L} 
\end{align*}
Note that for convolutional weights, $m_{T,l}$ is the number of nonzero parameters in $\bm{W}_{T}^{(l)} \in \mathcal{R}^{c_l \times c_{l-1} \times k_l}$.
The following width overparametrization of the output channels in a convolutional network
\begin{align*}
    q_{l-1} \geq \frac{\log\left(\frac{\delta}{m_{T,l}q_{l}L}\right)}{\log(1 - p_l)} = \frac{1}{\log (1 / (1 - p_l))} \log \left( \frac{Lm_{T,l}q_l}{\delta}\right)
\end{align*}
allows an ER network to represent the target with probability $1-\delta$.

The weights in the convolutional network can now be chosen as:

\begin{align*}
    w_{S, i' j' k}^{(l)} = w_{T, ijk}^{(l)}, \text{ for every }  i' \in \{q_{l}i, q_{l}i + 1, .., q_{l}i + q_{l}\},
\end{align*}
where $j' \in \{q_{l-1}j, q_{l-1}j + 1, .., q_{l-1}j + q_{l-1}\}$ is chosen randomly among all non-masked connections of $i'$ to block $j$ and the remaining connections are pruned away or set to zero.
The biases are set as in the proof of Theorem~\ref{thm:existence}.

\subsection{Lower Bound on the Overparametrization of ER networks}

\begin{figure*}[h!]
    \centering
        \centering
        \includegraphics[height=6cm, width=0.4\textwidth]{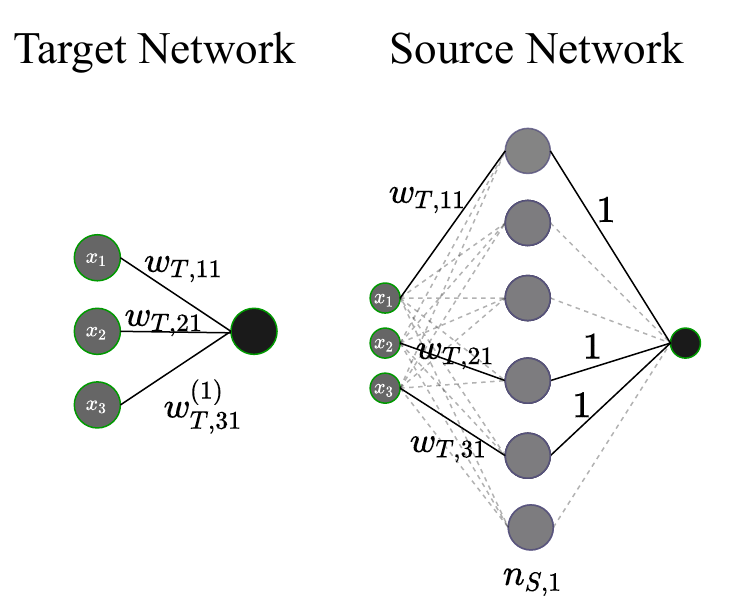}
    \caption{\textit{Lower bound} of width of an ER source network shown on the right required to represent the target network on the left. 
    The solid edges in the source on the right are the nonzero (unmasked) edges while the dotted lines are masked away in an ER source network.}
    \label{fig:lower_bound}
\end{figure*}

\label{app:lower-bound}
Our theoretical analysis suggests that ER networks require a width overparametrization by a factor of $\log(1 / \text{sparsity})$ to approximate an arbitrary target.
We also show that we cannot do substantially better than a width that is proportional to $\log(1 / \text{sparsity})$. 
\begin{theorem}[Lower bound on overparametrization in ER networks]
There exist univariate target networks $f_T(\bm{x}) = \phi(\bm{w}^T_T \bm{x} + b_T)$ that cannot be represented by a random $1$-hidden-layer ER source network $f_{S} \in \text{ER}(p)$ with probability at least $1-\delta$, if its width is $n_{S,1} < \frac{1}{\log(1 / (1-p))} \log\left( \frac{1}{1-(1-\delta)^{1/d}}\right)$.
\end{theorem}

\textbf{Proof}: The main idea is to find the minimum width of a single hidden layer network $ER(p)$ which can approximate a single output target $f_T(\bm{x}) = \phi(\bm{w}^T_T \bm{x} + b_T)$.
This minimum would be achieved when every target weight in $\bm{w}_T$ is approximated by exactly one path in the ER network from the input to the output (through the hidden layer).
We derive the probability that for every weight in the target, there is at least one non-masked path in the ER source that can represent this weight as shown in Figure \ref{fig:lower_bound}.
Bounding this probability will give us a lower bound on the minimum width required in the ER network to be able to represent the target network.
There are $n_{S,1}$ paths from an input neuron to an output neuron in the source network and the probability that each of this path exists is $p^2$, independently for each path, since both the input and output links in the path must be nonzero and each edge exists independently.
Starting from the first input neuron, the probability that there is at least one path from input $x_i$ to the output is $\left(1 - (1 - p^2)^{n_{S,1}}\right)$.
The paths exist independelty from each other if they start in different input neurons.
Thus, the probability that we can represent an arbitrary target neuron with $d$ input neurons is $\left(1 - (1 - p^2)^{n_{S,1}}\right)^d$.
In order to find the minimum width required, we lower bound this probability as:
\begin{align*}
   \left(1 - (1 - p^2)^{n_{S,1}}\right)^d \geq 1 - \delta
\end{align*}
Solving this inequality for $n_{S,1}$ proves the statement, since we would need
\begin{align*}
    n_{S,1} \geq \frac{1}{\log(1 / (1 - p^2))}\log \left(\frac{1}{(1 - (1 - \delta)^{1/d})}\right)
    \geq \frac{1}{\log(1 / (1 - p))}\log \left(\frac{1}{(1 - (1 - \delta)^{1/d})}\right).
\end{align*}

\subsection{Experimental Setup}
\label{app:expt-setup}
We conduct our experiments with two datasets built for image classification tasks: CIFAR10 and CIFAR100 \cite{cifar10-100}. 
Experiments on Tiny Imagenet \citep{tiny-imagenet} are reported in Appendix \ref{app:tiny-imagenet}.
We train two popular architectures, VGG16 \cite{vgg} and ResNet18 \cite{resnet}, to classify images in the CIFAR10 dataset. 
On the larger CIFAR100 dataset, we use VGG19 and ResNet50. 
Our code builds on the work of \cite{random-pruning-vita,synflow,kusupati2020soft} and is available at \url{https://github.com/RelationalML/sparse_to_sparse}.
All our experiments were run with 4 Nvidia A100 GPUs.

\textbf{Random Pruning}.
Each model is trained using Stochastic Gradient Descent (SGD) with learning rate $0.1$ and momentum $0.9$ with weight decay $0.0005$ and batch size $128$. 
We use the same hyperparameters as \cite{sanity-check-jackpot} and train every model for $160$ epochs. 
We repeat all our experiments over three runs and report averages and standard $0.95$-confidence intervals, which can be found in the appendix due to space constraints.
 
\textbf{Strong Lottery Tickets.} For experiments on strong lottery tickets using edge popup, we use an iterative version of edge popup as described in \citep{planting-jonas}.
We initialize a sparse network and anneal the sparsity iteratively while keeping the mask fixed. For the ResNet18 we use a learning rate of $0.1$ and anneal in $5$ levels and $100$ epochs for each level. The batch size is $128$ and we use SGD with momentum $0.9$ and weight decay $0.0005$. We report performances after one run for each of these experiments due to limited computation.

\textbf{Dynamic Sparse Training.} In the DST experiments, we use the same setup as random pruning, and modify the mask every $100$ iterations. For sparse to sparse training with DST, we us weight magnitude as importance score for pruning (with prune rate 0.5) and gradient for growth. 

\textbf{Sparse to Sparse Training.}
For the baseline IMP, we prune the network by removing $20\%$ parameters in every cycle and training for $150$ epochs in each cycle with a learning rate $0.1$ and a cosine LR schedule that anneals the learning rate to $0.01$.
We follow the same procedure while training an ER network.

For continuous sparsification with STR, we use the code provided by the authors \citep{kusupati2020soft} and the same hyperparameters
for both ResNet18 and ResNet50 with \texttt{sInit\_value}$ = -200$ and modify the weight decay parameter as per the target sparsity. $0.0005$ for target sparsity $0.96$ and $0.001$ for target sparsity $0.995$.

\subsection{Additional expriments  on CIFAR10/100 for Performance of Random Pruning}
\label{app:wlt-resnet-cifar}
Along with VGG we report results for different layerwise sparsity ratios for ResNets.
We use a ResNet 18 for CIFAR10 and a ResNet 50 for CIFAR100.

\begin{table}[H]
\centering
\begin{tabular}{c |  c  c  c  c } 
Sparsity & Pyramidal &  Balanced  & Uniform & ERK  \\
\midrule 
$0.9$ & $94.17 \pm 0.1$ & $93.97 \pm 0.13$ & $92.85 \pm 0.2$ & $93.96 \pm 0.19$ \\
$0.99$ & $\bm{90.83 \pm 0.3}$ & $90.72 \pm 0.3$ & $84.7 \pm 0.2$ & $89.04 \pm 0.21$ \\
$0.995$ & $\bm{88.57 \pm 0.12}$ & $88.32 \pm 0.3$ & $77.2 \pm 1$ & $85.8 \pm 0.16$\\
$0.999$ & $10 \pm 0$ & $70.69 \pm 0.6$ & $35.31 \pm 4$ & $61.36 \pm 0.24$ \\
\bottomrule
\end{tabular}%
\vspace{0.1cm}
\begin{tabular}{c |  c  c  c  c  } 
Sparsity &  Snip (ER) & Synflow (ER) & IMP (ER) \\
\midrule 
$0.9$ &  $\bm{94.25 \pm 0.3}$ & $93.95 \pm 0.13$ & $93.36 \pm 0.2$\\
$0.99$ & $90.33 \pm 0.04$  & $91.34 \pm 0.27$ & $86.43 \pm 0.4$\\
$0.995$ & $87.72\pm 0.14$ & $88.64 \pm 0.14$ & $ 81.2 \pm 0.16$\\
$0.999$ & $10 \pm 0$ & $\bm{71.92 \pm 0.28}$ & $50.46 \pm 1.4$\\
\bottomrule
\end{tabular}%
\vspace{0.1cm}
\caption{\textit{ER networks with different layerwise sparsities on CIFAR10 with ResNet18.}}
\label{table:results-res-cifar10}
\end{table}

\begin{table}[H]
\centering
\begin{tabular}{c |  c  c  c  c c  } 

Sparsity & Pyramidal &  Balanced  & Uniform & ERK & Snip (ER)  \\
\midrule 
$0.5$ & $78.09 \pm 0.64$ & $76.98 \pm 0.55$ & $\bm{78.12 \pm 0.33}$ & $77.63 \pm 0.52$ & $78.02 \pm 0.43$ \\
$0.8$ & $\bm{78.44 \pm 0.41}$ & $76.59 \pm 0.33$ & $77.77 \pm 0.43$ & $77.08 \pm 0.44$ & $76.21 \pm 0.77$ \\
$0.9$ & $\bm{76.66 \pm 0.01}$ & $75.37 \pm 0.77$ & $75.94 \pm 0.27$ & $76.02 \pm 0.62$ & $76.35 \pm 0.32$ \\
$0.99$ & $65.44 \pm 1.2$ & $\bm{67.97 \pm 0.23}$ & $55.52 \pm 2.5$ & $65.52 \pm 0.3$ & $1 \pm 0$ \\
\bottomrule
\end{tabular}%
\vspace{0.1cm}
\caption{\textit{ER networks with different layerwise sparsities on CIFAR100 with ResNet50.}}
\label{table:results-res-cifar100}
\end{table}

\begin{table}[H]
\centering
\begin{tabular}{c |  c  c  c  c } 
Sparsity & Pyramidal &  Balanced  & Uniform & ERK  \\
\midrule 
$0.9$ & $92.92 \pm 0.31$ & $93.22 \pm 0.28$ & $91.31 \pm 0.3$& $92.72 \pm 0.46$ \\
$0.99$ & $\bm{90.41 \pm 0.03}$ & $89.31 \pm 0.1$& $82.68 \pm 0.21$ & $87.81 \pm 0.38$ \\
$0.995$ & $87.76 \pm 0.13$ & $\bm{85.92 \pm 0.4}$&$73.69 \pm 0.64$ & $84.53 \pm 0.2$ \\
$0.999$ & $10 \pm 0$ & $\bm{68.68}$& $14.24$& $59.22 \pm 2.6$ \\
\bottomrule
\end{tabular}%
\vspace{0.1cm}
\begin{tabular}{c |  c  c  c  } 
Sparsity &  Snip (ER) & Synflow (ER) & IMP (ER) \\
\midrule 
$0.9$ &  $\bm{93.23 \pm 0.2}$ & $91.4 \pm 0.11$& $90.05 \pm 0.3$\\
$0.99$ &  $26.32 \pm 28$ & $86.55 \pm 0.26$& $90.15 \pm 0.05$\\
$0.995$ &  $10 \pm 0$ & $84.03 \pm 0.06$& $79.02 \pm 8$\\
$0.999$  & $10 \pm 0$ & $63.81 \pm 1$ & $10 \pm 0$\\
\bottomrule
\end{tabular}%
\caption{\textit{ER networks with different layerwise sparsities on CIFAR10 with VGG16.} We compare our layerwise sparsity ratios \textit{balanced} and \textit{pyramidal} with the uniform baseline, ERK and ER networks with layerwise sparsitiy ratios obtained by IMP, Iterative Synflow and Snip.}
\end{table}

\begin{table}[H]
\centering
\begin{tabular}{c |  c  c  c  c } 
Sparsity & Pyramidal &  Balanced  & Uniform & ERK  \\
\midrule 
$0.5$ & $73.63 \pm 0.1$ & $\bm{73.92 \pm 0.23}$ & $72.77 \pm 0.01$ & $73.58 \pm 0.18$ \\
$0.8$ & $73.65 \pm 0.43$ & $73.51 \pm 0.25$ & $71.39 \pm 0.07$ & $72.82 \pm 0.36$ \\
$0.9$ & $72.73 \pm 0.5$ & $72.49 \pm 0.43$ & $69.06 \pm 0.35$ & $71.9 \pm 0.06$ \\
$0.99$ & $60.05 \pm 3.2$ & $\bm{65.33 \pm 0.35}$ & $55.79 \pm 0.22$ & $63.83 \pm 0.41$ \\
\bottomrule
\end{tabular}%
\vspace{0.1cm}
\begin{tabular}{c |  c c c } 
Sparsity &  Snip (ER) & Synflow (ER) & IMP (ER) \\
\midrule 
$0.5$ &  $73.75 \pm 0.57$ & $72.6 \pm 0.6$ & $71.03 \pm 0.4$\\
$0.8$ &   $\bm{74.01 \pm 0.02}$ & $71.56 \pm 0.33$ & $68.09 \pm 0.14$\\
$0.9$ &  $\bm{73.05 \pm 0.24}$ & $70.8 \pm 0.31$ & $62.97 \pm 0.7$\\
$0.99$ & $1 \pm 0$ & $62.62 \pm 0.12$ & $1 \pm 0$\\
\bottomrule
\end{tabular}%
\vspace{0.1cm}
\caption{\textit{ER networks with different layerwise sparsities on CIFAR100 with VGG19.} We compare our layerwise sparsity ratios \textit{balanced} and \textit{pyramidal} with the uniform baseline, ERK and ER networks with layerwise sparsitiy ratios obtained by IMP, Iterative Synflow and Snip.}
\end{table}

\subsection{Dense Training Baselines on CIFAR10}
\label{app:dense-sparse-baseline}
For reference, we provide the baselines of STR and IMP on CIFAR10 with a ResNet 18 starting from a dense network to achieve the target sparsity in Tables \ref{table:str-baseline-c10}, \ref{table:imp-baseline-c10}.
\begin{table}[h!]
\centering
\begin{tabular}{ c | c c} 

Final Sparsity  &   $0.9$ &  $0.99$ \\
 \midrule
 
 Balanced &$93.38 \pm 0.12$ &$91.39 \pm 0.4$ \\
 
\end{tabular}
     \caption{\textit{IMP baseline on CIFAR10 for ResNet18}.} 
     \label{table:imp-baseline-c10}
\end{table}

\begin{table}[h!]
\centering
\begin{tabular}{ c | c c} 

Final Sparsity  &   $0.9$ &  $0.993$ \\
 \midrule
 
 Balanced &$94.66 \pm 0.09$ &$90.95 \pm 0.08$ \\
 
\end{tabular}
     \caption{\textit{STR baseline on CIFAR10 for ResNet18}.} 
     \label{table:str-baseline-c10}
\end{table}

\subsection{Sparse to Sparse Training}
We provide additional experiments for Sparse to Sparse training with a ResNet50 on CIFAR100 in Table \ref{table:app-str-er-cifar100}.
We also report the confidence intervals for the results in the main paper in Table \ref{table:str-er}, \ref{table:imp-er}.
\label{app:sparse-sparse}
\begin{table}[h!]
\centering
\begin{tabular}{l | c c c c}
Initial Sparsity & $0.7$ &$0.7$ &$0.8$ &$0.9$  \\
\midrule
Final Sparsity& $0.96$ & $0.997$ & $0.997$ & $0.998$\\
\midrule
Balanced &$94.11 \pm 0.07$ &$\mathbf{90.7 \pm 0.25}$ & $\mathbf{90.28 \pm 0.08}$& $\mathbf{89.47 \pm 0.2}$\\
Pyramidal &$94.18 \pm 0.12$ &$90.1 \pm 0.2$ & $89.52 \pm 0.18$& $88.87 \pm 0.34$\\
ERK &$\mathbf{94.33 \pm 0.28}$ &$90.12 \pm 0.2$ & $89.51 \pm 0.12$&$88.25 \pm 0.19$\\
Uniform &$93.74 \pm 0.16$ &$88.92 \pm 0.11$ &$87.89 \pm 0.22$ &$86.07 \pm 0.36$\\
STR (ER) &$93.89 \pm 0.19$ &$89.31 \pm 0.53$ &$87.87 \pm 0.19$ &$85.86 \pm 0.61$\\

\end{tabular}
\caption{\textit{Sparse to sparse training with Soft Threshold Reparametrization in ER networks:} Results on a ResNet18 trained on CIFAR10.}
\label{table:app-str-er}
\end{table}

\begin{table}[h!]
\centering
\begin{tabular}{l | c c c c }
Initial Sparsity & $0.7$ &$0.7$ & $0.8$ & $0.9$\\
\midrule
Final Sparsity & $0.9$ & $0.99$ & $0.93$ &$0.97$ \\
\midrule
Balanced &$93.54 \pm 0.12$  &$90.72 \pm 0.02$ &$93.14 \pm 0.22$ &$91.89 \pm 0.19$ \\
Pyramidal &$\mathbf{93.65 \pm 0.04}$  &$\mathbf{92.23 \pm 0.0.36}$ &$93.24 \pm 0.1$ &$92.23 \pm 0.4$ \\
ERK &$93.5 \pm 0.01$  &$90.95 \pm 0.12$ &$\mathbf{93.57 \pm 0.53}$ &$\mathbf{93.21 \pm 0.23}$ \\
Uniform &$93.18 \pm 0.005$  &$90.15 \pm 0.03$ &$92.62 \pm 0.07$ &$90.41 \pm 0.24$ \\
 
\end{tabular}
\caption{\textit{Sparse to sparse training with Iterative Magnitude Pruning in ER networks:} Results on a ResNet18 trained on CIFAR10.}
\label{table:app-imp-er}
\end{table}

\begin{table}[h!]
\centering
\begin{tabular}{l | c c c c}
Initial Sparsity & $0.7$  \\
\midrule
Final Sparsity& $0.995$ \\
\midrule
Balanced &$ 67.29 \pm 0.13$ \\
Pyramidal &$ 67.88\pm 0.34$ \\
ERK &$\mathbf{ 67.67\pm 0.45}$ \\
Uniform &$ 66.51\pm 0.4$ \\
STR (ER) &$ 66.8\pm 0.61$ \\
                    
\end{tabular}
\caption{\textit{Sparse to sparse training with Soft Threshold Reparametrization in ER networks:} Results on a ResNet50 trained on CIFAR100.}
\label{table:app-str-er-cifar100}
\end{table}

\subsection{Additional experiments for Strong Lottery Tickets in ER networks}\label{app:SLTexp}
To show experimentally that ER networks can contain SLTs, we use edge popup \cite{ramanujan-slth} to search for SLTs in ER ResNet18. 
We gradually anneal the sparsity of the ER network with 5 levels as proposed by \citep{planting-jonas}.
The results starting from ER networks with different initial sparsities are presented in Table \ref{table:slt_results_complete}. 
The confidence intervals of Table \ref{table:slt_results} are reported in Table \ref{table:app_slt_results}.
As a reference, we also report baseline results for dense networks in Table \ref{table:slt-ref}.

\begin{table}[h!]
\centering
\begin{tabular}{ c | c  c c  c } 
Initial Sparsity &$0.7$ &$0.5$ &$0.8$ &$0.5$ \\
\midrule
Final Sparsity  &   $0.9$ &  $0.95$ &  $0.95$ &  $0.99$  \\
 \midrule
 Uniform &  $88 \pm 0.3$  & $87.8 \pm 0.3$   & $\mathbf{88.1 \pm 0.1}$  & $87.9 \pm 0.1$\\
 Balanced &$\mathbf{88.06 \pm 0.13}$ &$87.93 \pm 0.38$ &$87.86 \pm 0.18$ &$87.93 \pm 0.14$ \\
 Pyramidal &$87.73 \pm 0.24$ &$\mathbf{88.02 \pm 0.03}$ &$87.95 \pm 0.14$ &$\mathbf{87.97 \pm 0.12}$  \\
 ERK &$88.04\pm0.13$ &$87.76\pm0.2$  &$88.02\pm0.2$ &$87.85\pm0.11$ \\
\end{tabular}
     \caption{\textit{ER networks for Strong Lottery Tickets}: Average results and $0.95$ standard confidence intervals for training an ER ResNet18 network with edge popup \citep{ramanujan-slth} on CIFAR10 across three runs. The ER network is gradually annealed to attain a SLT of the final sparsity (initial $\rightarrow$ final sparsity). Note that the first column serves as a baseline i.e. starting from a dense network.} 
     \label{table:app_slt_results}
\end{table}

\begin{table}[H]
\centering

\begin{tabular}{ c | c | c c  } 
Sparsity  & $0.5 \rightarrow 0.8$ &   $0.5 \rightarrow 0.9$ &   $0.7 \rightarrow 0.9$  \\
 \midrule
 Test Acc. & $87.83 \pm 0.25$ & $88.12 \pm 0.29$ & $87.95 \pm 0.25$  \\
\end{tabular}
\vspace{1 em}

\begin{tabular}{ c | c c | c}
    Sparsity &  $0.5 \rightarrow 0.95$ &  $0.8 \rightarrow 0.95$ &  $0.5 \rightarrow 0.99$  \\
    \midrule
    Test Acc. & $87.78 \pm 0.33$   & $88.07 \pm 0.06$  & $87.94 \pm 0.14$ \\
\end{tabular}
\vspace{0.1cm}
     \caption{\textit{ER networks for SLTs and different final sparsities}: Average results on training an ER ResNet18 network with edge popup \citep{ramanujan-slth} on CIFAR10. The ER network is initialized with a uniform initial sparsity, which is gradually annealed to attain a SLT of the final sparsity (initial $\rightarrow$ final sparsity). 
     Baseline results for initially dense networks are reported in Table~\ref{table:slt-ref}.}
     \label{table:slt_results_complete}
\end{table}

\begin{table}[H]
\centering
\begin{tabular}{ r |  c  r  c r} 
 Final Sparsity  & $0 \rightarrow 0.8$ & $0 \rightarrow0.9$ & $0 \rightarrow0.95$ & $0 \rightarrow0.99$  \\
 \midrule
\centering
Test Acc. & $87.79 \pm 0.1$ & $87.86\pm 0.2$ & $88 \pm 0.3$ & $87.7 \pm 0.56$\\
 
\end{tabular}
\caption{\textit{Baseline for edge popup with ResNet18 on CIFAR10}: The results for finding a SLT using edge popup starting from a dense network are shown. Our ER results starting from a sparse network are comparable to these baseline results which validates the efficiency of ER networks.}
\label{table:slt-ref}
\end{table}

Additional experiments for ER VGG16 on CIFAR10 are shown in Table \ref{table:slt_vgg} with baseline results in Table \ref{table:slt-ref-vgg} for ER networks with uniform sparsity.
\begin{table}[h!]
\centering
\begin{tabular}{ c | c c c c } 
Sparsity  & $0.5 \rightarrow 0.8$ &   $0.5 \rightarrow 0.9$ &   $0.5 \rightarrow 0.95$ &  $0.5 \rightarrow 0.99$\\
 \midrule
 Test Acc. & $88.03 \pm 0.26$ & $88.31 \pm 0.29$ & $88.06 \pm 0.35$  & $88.12 \pm 0.2$ \\
\end{tabular}
\vspace{0.1cm}
     \caption{\textit{ER networks for Strong Lottery Tickets}: SLTs in VGG16 ER networks on CIFAR10. The ER network is initialized with a uniform initial sparsity and gradually annealed to attain a SLT of the final sparsity (initial → final sparsity).}
     \label{table:slt_vgg}
\end{table}

\begin{table}[H]
\centering
\begin{tabular}{ r |  c  r  c r} 
 Final Sparsity  & $0 \rightarrow0.8$ & $0 \rightarrow0.9$ & $0 \rightarrow0.95$ & $0 \rightarrow0.99$  \\
 \midrule
\centering
Test Acc. & $88.2 \pm 0.15$ & $88.38 \pm 0.14$ & $88.16 \pm 0.21$ & $88.14 \pm 0.35$\\
 
\end{tabular}
\caption{\textit{Baseline for edge popup with VGG16 on CIFAR10}: Baseline results of Edge Popup to obtain SLTs on CIFAR10 with VGG16.}
\label{table:slt-ref-vgg}
\end{table}

\subsubsection{SLTs in ER networks for ResNet110 on CIFAR100}
We find SLTs within ER networks using the Edge Popup algorithm for a larger Resnet110 model as well, as reported in Table \ref{table:resnet110-slt}, for ER networks starting with uniform sparsity.
\begin{table}[h!]
\centering
\begin{tabular}{l | c c c}
\multirow{2}{*}{ER Method} & \multicolumn{3}{c}{Sparsity}
\\ 
\cmidrule{2-4}
 &$0 \rightarrow 0.9$ & $0.5 \rightarrow 0.9$ & $0.7 \rightarrow 0.9$ \\ \midrule
Test Acc.  & $61.91 \pm 0.13$ & $61.76 \pm 0.53$  & $61.78 \pm 0.61$  \\

\end{tabular}
\caption{Edge popup (SLTs) results on ER networks with uniform sparsity of Resnet110 on CIFAR100. Results are reported for one run due to limited compute.}
\label{table:resnet110-slt}
\end{table}

\subsection{Scalability of Random Pruning with Resnet110 on CIFAR100}

To showcase the scaleability of the suggested algorithms, we additionally report experiments for a larger model, i.e., Resnet110.

\subsubsection{ER networks for Random Pruning}

We report results for different layerwise sparsities in Table \ref{table:resnet110-wlt}.
As a reference we also report results on pruning with a baseline algorithm Iterative Magnitude Pruning in Table \ref{table:resnet110-imp}.
We also conducted experiments with an Iterative Synflow algorithm \citet{synflow} to prune a Resnet110 but the algorithm fails for such a large model.

\begin{table}[H]
\centering
\begin{tabular}{l | c c c c}
\multirow{2}{*}{ER Method}& \multicolumn{3}{c}{ Sparsity }                             \\ \cmidrule{2-5} 
 & $0.5$ &   $0.8$ &   $0.9$ & $0.95$ \\ \midrule
Balanced (ours) & $ 70.37 \pm 0.59$  & $67.88 \pm 1.01$ & $67.31 \pm 0.33$ & $63.80 \pm 0.09$ \\
Pyramidal (ours)  & $\mathbf{71.16 \pm 0.22}$ & $69.56 \pm 0.31$  & $63.23 \pm 1.29$ & $52.37 \pm 0.51$\\
ERK & $70.76 \pm 0.82$  & $\mathbf{69.96 \pm 0.58}$  & $\mathbf{68.14 \pm 0.34}$ & $\mathbf{64.92 \pm 0.31}$ \\
Uniform & $70.86 \pm 0.70$  & $69.41 \pm 0.27$  & $66.32 \pm 0.42$ & $61.31 \pm 0.02$\\
ER Snip & $69.14 \pm 0.46$  & $69.12 \pm 0.65$  & $65.82 \pm 0.28$ & $60.15 \pm 0.34$\\

\end{tabular}
\caption{\textit{Results for random pruning on CIFAR100 with ResNet110}. Results are average and standard deviation reported across three runs.}
\label{table:resnet110-wlt}
\end{table}


\begin{table}[H]
\centering
\begin{tabular}{l | c c}
\multirow{2}{*}{Iterative Magnitude Pruning} & \multicolumn{2}{c}{Sparsity}
\\ 
\cmidrule{2-3}
 &$0.5$ &  $0.9$ \\ 
 \midrule
Test Acc.  & $65.46$ &  $64.77$  \\

\end{tabular}
\caption{Results on CIFAR100 with Resnet110 pruning with the iterative magnitude pruning (IMP) algorithm for reference. Only one run of IMP was performed for each of these sparsities.}
\label{table:resnet110-imp}
\end{table}

\subsection{Experiments on Random Pruning with Tiny Imagenet}

We also report experiments with different layerwise sparsity ratios in ER networks for the Tiny Imagenet dataset. We use a VGG19 and a ResNet20 and show that our proposed layerwise sparsity methods for ER networks are competetive for this dataset. 
Note that we use the validation set provided by the creators of Tiny Imagenet \citep{tiny-imagenet} as a test set to measure the generalization performance of our trained models.

See Table \ref{table:tiny-img-vgg} and \ref{table:tiny-img-resnet}.

\label{app:tiny-imagenet}

\begin{table}[H]
\centering
\begin{tabular}{l | c c c c}
\multirow{2}{*}{ER Method}& \multicolumn{3}{c}{ Sparsity }                             \\ \cmidrule{2-5} 
 & $0.5$ &   $0.8$ &   $0.9$ & $0.99$ \\ \midrule
Balanced (ours) & $ 58.40 \pm 0.39$  & $57.95 \pm 0.42$ & $57.47 \pm 0.64$ & $\mathbf{50.72 \pm 0.15}$ \\
Pyramidal (ours)  & $\mathbf{58.92 \pm 0.12}$ & $58.46 \pm 0.15$  & $\mathbf{58.08 \pm 0.05}$ & $41.06 \pm 0.28$\\
ERK & $58.66 \pm 0.63$  & $58.39 \pm 0.15$  & $57.24 \pm 0.12$ & $50.52 \pm 0.50$\\
Uniform & $58.67 \pm 0.27$  & $57.61 \pm 0.36$  & $54.96 \pm 0.82$ & $44.78 \pm 1.14$\\
ER Snip & $58.66 \pm 0.29$  & $\mathbf{58.65 \pm 0.29}$  & $57.75 \pm 0.18$ & $0.5$\\

\end{tabular}
\caption{\textit{Results for ER networks on Tiny Imagenet with VGG19}}
\label{table:tiny-img-vgg}
\end{table}

\begin{table}[H]
\centering
\begin{tabular}{l | c c c c}
\multirow{2}{*}{ER Method}& \multicolumn{3}{c}{ Sparsity }                             \\ \cmidrule{2-5} 
 & $0.5$ &   $0.8$ &   $0.9$ & $0.99$ \\ \midrule
Balanced (ours) & $ 46.92 \pm 0.38$  & $39.25 \pm 0.57$ & $31.62 \pm 0.49$ & $9.34 \pm 0.46$ \\
Pyramidal (ours)  & $48.24 \pm 0.01$ & $39.31 \pm 0.17$  & $25.15 \pm 0.08$ & $1.66 \pm 0.13$\\
ERK & $\mathbf{50.36 \pm 0.47}$  & $\mathbf{44.73 \pm 0.64}$  & $36.81 \pm 0.32$ & $\mathbf{10.52 \pm 0.05}$ \\
Uniform & $48.49 \pm 0.47$  & $41.84 \pm 0.38$  & $32.87 \pm 0.15$ & $8.70 \pm 1.14$\\
ER Snip & $50.28 \pm 0.50$  & $44.65 \pm 0.47$  & $\mathbf{36.92 \pm 0.20}$ & $7.36 \pm 1.81$\\

\end{tabular}
\caption{\textit{Results for ER networks on Tiny Imagenet with ResNet20}}
\label{table:tiny-img-resnet}
\end{table}

\subsection{Dynamic Sparse Training on ER networks}
\label{app:dst}

\begin{table}[H]
\centering
\begin{tabular}{ l   l  c  l c } 
 & \multicolumn{2}{c}{Sparsity $0.99$} &  \multicolumn{2}{c}{Sparsity $0.995$}  \\
\cmidrule{2-5}
 ER Method  & Original & Rewired & Original & Rewired \\
 \midrule 
 ERK  & $87.81 \pm 0.39$ & $90.78 \pm 0.14$ & $84.53 \pm 0.20$& $88.28 \pm 0.52$\\ 
 Balanced & $89.31 \pm 0.11$ & $91.41 \pm 0.43$& $85.91 \pm 0.40$ & $89.30 \pm 0.03$\\ 
 Pyramidal & $\mathbf{90.41 \pm 0.03} $ & $\mathbf{91.97 \pm 0.08}$& $\mathbf{87.76 \pm 0.13}$ & $\mathbf{90.61 \pm 0.15}$\\ 
\end{tabular}%
\vspace{0.1cm}
\begin{tabular}{ l  l c} 
 &  \multicolumn{2}{c}{Sparsity $0.999$} \\
\cmidrule{2-3}
 ER Method  & Original & Rewired \\
 \midrule 
 ERK  &  $59.22 \pm 2.57$ & $74.12 \pm 1.16$\\ 
 Balanced & $\mathbf{68.68 \pm 0.44}$ & $\mathbf{78.90 \pm 0.64}$\\ 
 Pyramidal & $10 \pm 0$& $9.83 \pm 0.28$\\ 
\end{tabular}%
\vspace{0.1cm}
\caption{\textit{ER networks rewired with DST:} An $ER(\mathbf{p})$ VGG16 network with sparsity = $1-p$ is initialized and the mask is modified by rewiring edges with RiGL on CIFAR10.} 
\label{app:const_results}
\end{table}

In addition to the rewiring experiments shown in Table \ref{table:const_results}, we use Dynamical Sparse Training to prune an already sparse ER network to a higher sparsity and see if this can achieve the same performance as performing DST starting from a denser network.
Similar experiments have been shown by \citep{granet}.
However, we report results on  ER networks starting at much higher sparsities. 
Our results shown in Table \ref{table:granet} are able to match the performance of \citep{granet} while being more efficient as we start at a higher sparsity.
\begin{table}[h!]
\centering
\begin{tabular}{l | c c c}
\multirow{2}{*}{ER Method} & \multicolumn{3}{c}{Sparsity}
\\ 
\cmidrule{2-4}
 &$0.5 \rightarrow 0.99$ & $0.9 \rightarrow 0.99$ & $0.95 \rightarrow 0.99$ \\ \midrule
Balanced (ours)  & $93.08 \pm 0.01$ & $92.75 \pm 0.25$  & $\mathbf{92.70 \pm 0.10}$  \\
Pyramidal (ours) & $\mathbf{93.13 \pm 0.05}$  & $\mathbf{92.93 \pm 0.08}$  & $92.58 \pm 0.21$\\
ERK       & $92.94 \pm 0.12$  & $92.77 \pm 0.01$ & $92.47 \pm 0.11$ 
\end{tabular}
\caption{\textit{Sparse to sparse training with DST} Final test accuracy for VGG16 on CIFAR10 is reported where the model is initialized with an ER network of some initial sparsity and further pruned to a final sparsity (initial $\rightarrow$ final) while modifying the mask using the RiGL \citep{evci-rigl} algorithm.}
\label{table:granet}
\end{table}

Notably, we observe that it is also possible to start at a sparsity of up to $0.95$ and still achieve a competitive test accuracy, only marginally worse than starting with a sparsity of $0.5$.

\subsection{Visualizing layerwise sparsities for ER networks}\label{app:sparratio}

\begin{figure}[h!]
    \centering
    
    \includegraphics[width=\textwidth]{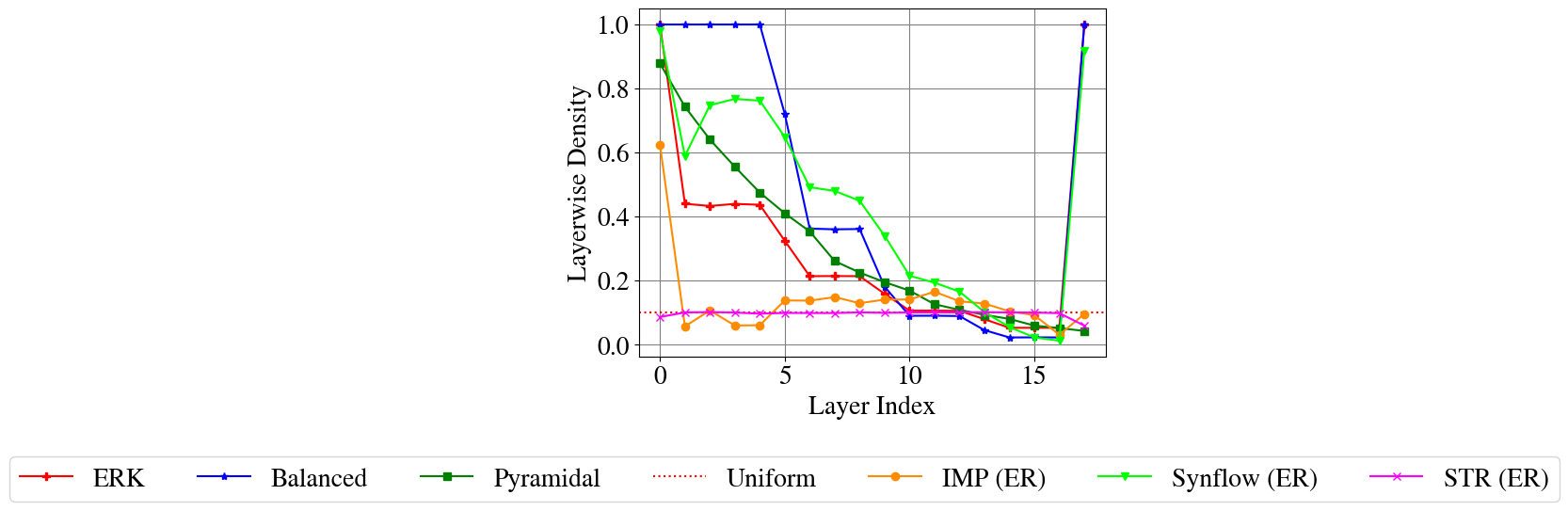}
    
    \caption{\textit{Layerwise density for ResNet18 trained on CIFAR10 for results reported in Table~\ref{table:results-res-cifar10}, \ref{table:app-str-er}} for target sparsity $0.9$ i.e. $10\%$ of the parameters are retained.}
    \label{fig:layerwise-c10}
\end{figure}

We report layerwise sparsity ratios for the proposed methods discussed in Section 3 in comparison to ERK and Uniform in Table \ref{fig:layerwise-c10}.

\subsection{Additional Experiments on Diverse Datasets}
\label{app:expt-additional}
To showcase the versatility of our insights, we present additional experiments on more diverse datasets with varying application domains. This verifies the applicability of sparse to sparse training in a broad context.

\textbf{ImageNet Experiments}

Table~\ref{table:imagenet-results}) establishes that sparse to sparse training also works in the context of large scale data like ImageNet \citep{ILSVRC15}, on which we train a sparse ResNet 50 using Iterative Magnitude Pruning.
The sparse ER network is initialized with a balanced layerwise sparsity ratio.
We find that starting from an ER network of $50\%$ sparsity, IMP is still able to find an $80\%$ sparse network without loss of performance.
\begin{table}[h!]
\centering
\begin{tabular}{ c | c c} 
Initial Sparsity &$0$ (dense) &$0.5$ \\
\midrule
Final Sparsity  &   $0.8$ &  $0.8$ \\
 \midrule
 
 Accuracy &$71.57$ &$\mathbf{71.73}$ \\
 
\end{tabular}
     \caption{\textit{Sparse to sparse training with IMP on ImageNet}: Accuracy of a ResNet 50 trained on ImageNet.
     The sparsity ratios of the initial ER network were chosen as balanced.} 
     \label{table:imagenet-results}
\end{table}

\textbf{Graph Convolutional Networks}
We show that random pruning can enable sparse training in Graph Convolutional Networks (GCN)\citep{gcn} (see Table \ref{table:gcn-results}).
We adapt the experimental setup and hyperparameters provided by \url{https://github.com/meliketoy/graph-cnn.pytorch}.
Each layer in the initial random GCN has uniform sparsity.
For each training cycle in IMP, we increase the sparsity by $10\%$ till the target sparsity is achieved.
Starting sparse only marginally affects final performance.
\begin{table}[h!]
\centering
\begin{tabular}{ c | c c} 
Initial Sparsity &$0$ (dense) &$0.4$ \\
\midrule
Final Sparsity  &   $0.7$ &  $0.7$ \\
 \midrule
 
 Accuracy & $83.33 \pm 0.007$ &$\mathbf{81.9 \pm 0.005}$ \\
 
\end{tabular}
     \caption{\textit{Sparse to sparse training with IMP on GCNs}: We train a $2$ layer GCN on a node classification task on the CORA\citep{gcn-data} dataset averaged across $10$ runs.
     The dense $2$ layer GCN achieves $82.57 (\pm 0.005)\%$ accuracy.
     The sparsity ratios of the initial ER network are uniform.
     } 
     \label{table:gcn-results}
\end{table}

\textbf{Algorithmic Data}

We test our theory for MLPs on algorithmic data which has been studied in the context of grokking on the Toy Model described in Section 2 of \citet{grokking}. 
The main task is to learn addition through symbols, for which we adapt the original experimental setup (\url{https://github.com/ejmichaud/grokking-squared/blob/main/notebooks/erics-implementation.ipynb}) to employ sparse to sparse training (see Table \ref{table:grokking-results}).
Each layer of the MLP has uniform sparsity.
We increase the sparsity by $10\%$ in each training cycle till the target sparsity is achieved.
We observe that for each individual run, sparse to sparse training achieves the same performance as its dense to sparse counterpart.
We hypothesize that the toy model is heavily overparametrized and hence allows starting sparse.
However, each individual run is not consistent and can have a significantly different performance likely due to the quality of the randomly sampled training data.

\begin{table}[h!]
\centering
\begin{tabular}{ c | c c} 
Initial Sparsity &$0$ (dense) &$0.3$ \\
\midrule
Final Sparsity  &   $0.6$ &  $0.6$ \\
 \midrule
 
Accuracy & $82.14\pm 3.57$ &$82.14\pm 3.57$ \\
 
\end{tabular}
 \caption{\textit{Sparse to sparse training on algorithmic data with IMP}. We train a model to learn addition of two numbers symbolically as described in \citet{grokking}. 
 The encoder and decoder of the model used in \citet{grokking} are initialized with uniform sparsity ratios for sparse to sparse training. The average test accuracy and 0.95 confidence intervals are reported for 3 independent runs.}
 
 \label{table:grokking-results}
     
\end{table}

\textbf{Tabular Data}

Tabular data is often studied in the context of fairness \citep{fairness-data}, see also \url{https://github.com/socialfoundations/folktables}. We use a four layer MLP with 256 hidden units for binary classication of fairness (see Table \ref{table:tabular-results}). 
Each layer of the MLP has uniform sparsity.
We train the model with the Adam \citep{adam} optimizer and a learning rate of $0.01$ for $20$ epochs in each training cycle and increase the sparsity by $10\%$ in each training cycle till the target sparsity is achieved.

We find that starting sparse does not impact the final performance of the model.

\begin{table}[h!]
\centering
\begin{tabular}{ c | c c} 
Initial Sparsity &$0$ (dense) &$0.5$ \\
\midrule
Final Sparsity  &   $0.9$ &  $0.9$ \\
 \midrule
 
 Uniform & $\mathbf{82.4 \pm 0.06}$ &  $82.39 \pm 0.05$  \\
 
\end{tabular}
     \caption{\textit{Sparse to sparse training on tabular data with IMP}. The average test accuracy and 0.95 confidence intervals are reported for 3 independent runs.}
     \label{table:tabular-results}
\end{table}